\documentclass[final,twocolumn,5p,times]{elsarticle}

\usepackage{amsmath}
\usepackage{forest}
\usepackage{diagbox}
\usepackage{hyperref}
\usepackage{tabularx}
\usepackage{enumitem, ragged2e}
\usepackage{rotating}
\usepackage{amssymb}
\usepackage{amsfonts}
\usepackage{nicematrix}
\usepackage{pifont}
\usepackage{float}
\usepackage{multirow}
\usepackage{makecell, booktabs}
\usepackage{xcolor}
\usepackage{longtable}

\journal{Neurocomputing}

\begin{document}

\begin{frontmatter}



\title{Federated Continual Learning: Concepts, Challenges, and Solutions}

\author[label1]{Parisa Hamedi}
\ead{parisa.hamedi@unb.ca}
 
\author[label1]{Roozbeh Razavi-Far\corref{cor1}}
\ead{roozbeh.razavi-far@unb.ca}
 
\author[label2]{Ehsan Hallaji}
\ead{hallaji@uwindsor.ca}
 
 \affiliation[label1]{organization={Faculty of Computer Science, University of New Brunswick},
             addressline={550 Windsor Street},
             city={Fredericton},
             postcode={E3B 5A3},
             state={NB},
             country={Canada}}

 \affiliation[label2]{organization={Department of Electrical and Computer Engineering, University of Windsor},
             addressline={401 Sunset Ave},
             city={Windsor},
             postcode={N9B 3P4},
             state={ON},
             country={Canada}}

\cortext[cor1]{Corresponding author.}

\begin{abstract}
Federated Continual Learning (FCL) has emerged as a robust solution for collaborative model training in dynamic environments, where data samples are continuously generated and distributed across multiple devices. This survey provides a comprehensive review of FCL, focusing on key challenges such as heterogeneity, model stability, communication overhead, and privacy preservation. We explore various forms of heterogeneity and their impact on model performance. Solutions to non-IID data, resource-constrained platforms, and personalized learning are reviewed in an effort to show the complexities of handling heterogeneous data distributions. Next, we review techniques for ensuring model stability and avoiding catastrophic forgetting, which are critical in non-stationary environments. Privacy-preserving techniques are another aspect of FCL that have been reviewed in this work. This survey has integrated insights from federated learning and continual learning to present strategies for improving the efficacy and scalability of FCL systems, making it applicable to a wide range of real-world scenarios.
\end{abstract}




\end{frontmatter}



\section{Introduction}

The increasing number of interconnected devices and the growing demand for the use of intelligent systems in practice has sparked the interest into the concepts of distributed and incremental learning \cite{10444954, dai_addressing_2023}. A noticeable trend is Federated Learning (FL), which has established itself as an efficient mechanism to accomplish collaborative training of machine learning models across decentralized devices and resolve privacy and communication issues by keeping data where it belongs \cite{pmlr-v54-mcmahan17a, Hallaji2024}. On the other hand, Continual Learning (CL) aims to solve the problem of forcing models to learn and forget knowledge as the target data distribution changes, and, thus, it is relevant to non-static scenarios \cite{10444954, 9349197}. There is growing interest in the broader distribution and generation of information. Consequently, FCL has attracted significant attention from researchers for its ability to integrate the strengths of FL and CL, making it ideal for practical applications such as distributed and incremental learning.

FCL draws the foundation from the literature in FL and CL \cite{10423871}. FL addresses key challenges such as data and system heterogeneity, communication overhead, and privacy preservation in distributed systems. It enables global model training across diverse clients without centralizing data. However, it assumes static data distributions, making it less suited for dynamic environments where data streams continuously. While CL addresses issues with the presence of non-stationary data, a method that could deal with issues through techniques such as experience replay, regularization, or modification in architecture so as to ensure models stay time-robust and adapt to new incoming tasks without forgetting obtained knowledge in the previous state. FCL bridges between these domains by integrating FL concepts along with CL to effectively control evolving data \cite{9892815, 10208460}. For example, heterogeneity and privacy challenges brought about by FL are more complex in FCL because data streams dynamically, while solutions provided for incremental learning should be adapted to work under FL constraints such as non-IID data and limited communication \cite{CRIADO2022263}. These cross-domain interactions help us to better understand how existing research on FL and CL is fundamentally important in advancing FCL.

FCL is imperative in facilitating intelligent systems, where data is both distributed and non-stationary. The application of FCL ranges from personalized healthcare to real-time IoT analytics and edge-based autonomous systems \cite{PARAGLIOLA202316, BACCARELLI2022376}. Some of the unique challenges in FCL, such as addressing global and local catastrophic forgetting, heterogeneous client behavior, and scalable knowledge transfer across clients, remain barely explored in the literature. These challenges require new methodologies that effectively combine FL's focus on collaboration and privacy with CL's ability to adapt over time. Furthermore, practical issues such as dynamic client participation, resource constraints, and real-world non-IID data distributions make things even more complicated. Despite its promise, the area of FCL lacks a comprehensive overview of its challenges, methodologies, and applications. This gap calls for a dedicated survey to consolidate knowledge, identify open problems, solutions, and guide future research.

Table \ref{tab:recentSurveys} provides a comprehensive overview of recent studies in FCL, FL, and CL, focusing on how they address key challenges and propose corresponding solutions. For instance, \cite{10423871} categorizes FCL methods based on knowledge fusion mechanisms and centers on the issue of forgetting in both synchronous and asynchronous FCL settings. Similarly, \cite{wang2024federatedcontinuallearningedgeai} investigates FCL from the perspective of three key task characteristics, including federated class continual learning, federated domain continual learning, and federated task continual learning. Other works such as \cite{dai_addressing_2023, CRIADO2022263, 10752992} partially address FCL by exploring topics like online FL, CL within FL. For instance, \cite{dai_addressing_2023} examines the overlap between federated, transfer, and online learning, aligning with FCL principles. \cite{CRIADO2022263} addresses data heterogeneity in FL, highlighting challenges from non-IID data and concept drift, and links these issues to CL. Similarly, \cite{10752992} investigates catastrophic forgetting across domains, including its impact within FCL settings. While these surveys have contributed valuable insights on FCL, focusing on knowledge fusion, task-specific classifications, and Edge-AI contexts. However, they lack a comprehensive taxonomy of key challenges, such as heterogeneity, communication overhead, and model stability, and offer limited cross-domain insights from FL and CL. 

As shown in the table, all recent CL surveys discuss catastrophic forgetting, a core challenge in CL. However, \cite{10721446} and \cite{ijcai2023p743} also provide an in-depth analysis of concept drift, which is less commonly addressed. Furthermore, \cite{10444954} and \cite{9349197} highlight dilemmas in CL through statistical analyses, visualizations, and diagrams, offering valuable insights into practical limitations.

Privacy is a fundamental concern in FL, and all recent FL surveys acknowledge it. Notably, \cite{Hallaji2024, 10429780} deeply examine various attack vectors and defense mechanisms in FL systems. Heterogeneity is widely recognized as a fundamental challenge in FL. For instance, \cite{LI2024127906} specifically addresses data heterogeneity by examining issues such as class imbalance and missing classes. In a broader context, \cite{LIU2024128019} highlights heterogeneity as a critical obstacle in FL, analyzing its impact across data, model, and system levels. Additionally, \cite{10833754} offers an extensive overview of heterogeneity in FL, covering a wide range of factors including data, model, task, communication, and device heterogeneity. 

Finally, while model stability, communication overhead, and resource constraints are recognized as major challenges in FL, they remain underexplored in the literature.

This survey provides a structured analysis of FCL by examining the intersection of challenges from FL and CL. It identifies key challenges, reviews existing methodologies, and explores future research directions. The main contributions are as follows:
\begin{itemize}
    \item We systematically examine how the challenges of FL and CL contribute to FCL, providing researchers with a comprehensive understanding of the current state of the domain and identifying key future research directions.

    \item We identify and categorize the fundamental challenges in FCL, distinguishing between global and local issues, and trace their origins to specific challenges inherent in FL and CL. This approach enhances the clarity of the underlying factors contributing to FCL challenges, helping researchers and practitioners to understand how these challenges manifest and interact at both global and local levels. By uncovering the root causes, our analysis paves the way for more targeted solutions and effective strategies, ultimately improving the performance, scalability, and robustness of FCL systems.

    \item We review and analyze current studies in FCL, examining the challenges addressed in each work. Based on this, we categorize and evaluate the existing solutions to FCL challenges, offering a clearer understanding of how different approaches that tackle these issues and identifying areas for future development. 
\end{itemize}

\begin{table*}[h]
    \centering
    \caption{Exploring challenges in federated learning, continual learning, and federated continual learning. This table provides an overview of recent studies. Here L and G stand for local catastrophic forgetting, and global catastrophic forgetting, respectively.
    }
    \label{tab:recentSurveys}
    \resizebox{\textwidth}{!}{
    \begin{tabular}{l c c c c c c c c c c c c c c c c c c c c c c c}
        \toprule
        \multirow{2}{*}{\diagbox{Challenges}{ Studies}}& \multirow{2}{*}{Subcategory} & \multicolumn{6}{c}{FCL}&\multicolumn{7}{c}{CL}&\multicolumn{9}{c}{FL}\\
        \cmidrule(r){3-8}\cmidrule(r){9-15} \cmidrule(r){16-24}
        &&Ours& \cite{wang2024federatedcontinuallearningedgeai}&\cite{10423871}& \cite{dai_addressing_2023}&\cite{CRIADO2022263}&\cite{10752992} & \cite{10721446} & \cite{ijcai2024p924} & \cite{10444954} & \cite{mendez2023how} & \cite{ijcai2023p743} & \cite{9349197} & \cite{MAI202228} & \cite{10429780} & \cite{10856890} & \cite{10833754} &  \cite{10.1145/3678181}  & \cite{ALSHARIF2024101251} & \cite{LIU2024128019} & \cite{LI2024110663} & \cite{LI2024127906} & \cite{Hallaji2024}\\
        \midrule
        \multirow{3}{*}{Heterogeneity} &Statistical Heterogeneity &\checkmark &\checkmark&\checkmark&\checkmark&\checkmark&\checkmark&\ding{55}&\ding{55}&\ding{55}&\ding{55}&\ding{55}&\ding{55}&\ding{55}&\ding{55}&\checkmark&\checkmark&\ding{55}&\checkmark&\checkmark&\checkmark&\checkmark&\ding{55}  \\

        &Conceptual Heterogeneity&\checkmark&\ding{55}&\checkmark&\checkmark&\ding{55}&\ding{55}&\ding{55}&\ding{55}&\ding{55}&\ding{55}&\ding{55}&\ding{55}&\ding{55}&\ding{55}&\checkmark&\checkmark&\ding{55}&\checkmark&\ding{55}&\ding{55}&\checkmark&\ding{55}  \\

        &System Heterogeneity&\checkmark &\checkmark&\ding{55}&\checkmark&\ding{55}&\ding{55}&\ding{55}&\ding{55}&\ding{55}&\ding{55}&\ding{55}&\ding{55}&\ding{55}&\ding{55}&\checkmark&\checkmark&\ding{55}&\ding{55}&\checkmark&\ding{55}&\checkmark&\ding{55}  \\
        
        \midrule

        Resource Constraints& -- &\checkmark&\ding{55}&\ding{55}&\ding{55}&\ding{55}&\checkmark&\ding{55}&\ding{55}&\ding{55}&\ding{55}&\ding{55}&\ding{55}&\ding{55}&\ding{55}&\ding{55}&\checkmark&\ding{55}&\ding{55}&\ding{55}&\checkmark&\ding{55}&\checkmark  \\
        
        \midrule
        
        Communication Overhead & -- &\checkmark&\checkmark&\checkmark&\ding{55}&\ding{55}&\ding{55}&\ding{55}&\ding{55}&\ding{55}&\ding{55}&\ding{55}&\ding{55}&\ding{55}&\checkmark&\checkmark&\ding{55}&\checkmark&\checkmark&\ding{55}&\checkmark&\checkmark&\ding{55}  \\
        
        \midrule

        Model Stability & -- &\checkmark&\ding{55}&\ding{55}&\ding{55}&\ding{55}&\ding{55}&\ding{55}&\ding{55}&\ding{55}&\ding{55}&\ding{55}&\ding{55}&\ding{55}&\ding{55}&\checkmark&\ding{55}&\ding{55}&\ding{55}&\ding{55}&\ding{55}&\checkmark&\ding{55}  \\

        \midrule
        
        \multirow{4}{*}{Privacy Preservation} &Homomorphic Encryption&\checkmark&\ding{55}&\ding{55}&\ding{55}&\checkmark&\ding{55}&\ding{55}&\ding{55}&\ding{55}&\ding{55}&\ding{55}&\ding{55}&\ding{55}&\checkmark&\ding{55}&\checkmark&\checkmark&\checkmark&\checkmark&\checkmark&\ding{55}&\checkmark  \\

        &Differential Privacy&\checkmark&\checkmark&\ding{55}&\checkmark&\checkmark&\ding{55}&\ding{55}&\ding{55}&\ding{55}&\ding{55}&\ding{55}&\ding{55}&\ding{55}&\checkmark&\ding{55}&\checkmark&\checkmark&\checkmark&\checkmark&\checkmark&\checkmark&\checkmark  \\

        &Secure Multi-Party Computation &\checkmark&\ding{55}&\ding{55}&\ding{55}&\checkmark&\ding{55}&\ding{55}&\ding{55}&\ding{55}&\ding{55}&\ding{55}&\ding{55}&\ding{55}&\checkmark&\ding{55}&\checkmark&\checkmark&\checkmark&\ding{55}&\checkmark&\ding{55}&\checkmark  \\

        &Secure Aggregation&\checkmark&\ding{55}&\ding{55}&\ding{55}&\ding{55}&\ding{55}&\ding{55}&\ding{55}&\ding{55}&\ding{55}&\ding{55}&\ding{55}&\ding{55}&\ding{55}&\checkmark&\ding{55}&\checkmark&\checkmark&\checkmark&\ding{55}&\ding{55}&\checkmark \\
        
        \midrule
        
        \multirow{3}{*}{Concept Drift}&Virtual Drift&\checkmark&\checkmark&\checkmark&\checkmark&\checkmark&\checkmark&\checkmark&\ding{55}&\ding{55}&\ding{55}&\checkmark&\ding{55}&\ding{55}&\ding{55}&\ding{55}&\ding{55}&\ding{55}&\ding{55}&\ding{55}&\ding{55}&\ding{55}&\ding{55}\\
        
        &Real Drift&\checkmark&\ding{55}&\ding{55}&\ding{55}&\checkmark&\ding{55}&\ding{55}&\ding{55}&\ding{55}&\ding{55}&\checkmark&\ding{55}&\ding{55}&\ding{55}&\ding{55}&\ding{55}&\ding{55}&\ding{55}&\ding{55}&\ding{55}&\ding{55}&\ding{55}\\
        
        &Hybrid Drift&\checkmark&\ding{55}&\ding{55}&\checkmark&\checkmark&\ding{55}&\ding{55}&\ding{55}&\ding{55}&\ding{55}&\checkmark&\ding{55}&\ding{55}&\ding{55}&\ding{55}&\ding{55}&\ding{55}&\ding{55}&\ding{55}&\ding{55}&\ding{55}&\ding{55}\\
        \midrule
        Dilemmas& -- &\checkmark&\ding{55}&\checkmark&\ding{55}&\ding{55}&\ding{55}&\checkmark&\ding{55}&\checkmark&\ding{55}&\checkmark&\checkmark&\checkmark&\ding{55}&\ding{55}&\ding{55}&\ding{55}&\ding{55}&\ding{55}&\ding{55}&\ding{55}&\ding{55}\\
        \midrule
        \multirow{2}{*}{Catastrophic Forgetting}&Local Forgetting&\checkmark&\checkmark&\checkmark&\ding{55}&\checkmark&\checkmark&\checkmark&\checkmark&\checkmark&\checkmark&\checkmark&\checkmark&\checkmark&\ding{55}&\ding{55}&\ding{55}&\ding{55}&\ding{55}&\ding{55}&\ding{55}&\ding{55}&\ding{55}\\
        &Global Forgetting&\checkmark&\checkmark&\checkmark&\ding{55}&\checkmark&\checkmark&\ding{55}&\ding{55}&\ding{55}&\ding{55}&\ding{55}&\ding{55}&\ding{55}&\ding{55}&\ding{55}&\ding{55}&\ding{55}&\ding{55}&\ding{55}&\ding{55}&\ding{55}&\ding{55}\\
        
        \midrule
        Global Model Overfitting& -- &\checkmark&\ding{55}&\ding{55}&\ding{55}&\ding{55}&\ding{55}&\ding{55}&\ding{55}&\ding{55}&\ding{55}&\ding{55}&\ding{55}&\ding{55}&\ding{55}&\ding{55}&\ding{55}&\ding{55}&\ding{55}&\ding{55}&\ding{55}&\ding{55}&\ding{55}\\

        \midrule
        Negative Knowledge Transfer& -- &\checkmark&\checkmark&\ding{55}&\ding{55}&\ding{55}&\ding{55}&\ding{55}&\ding{55}&\ding{55}&\ding{55}&\ding{55}&\ding{55}&\ding{55}&\ding{55}&\ding{55}&\ding{55}&\ding{55}&\ding{55}&\ding{55}&\ding{55}&\ding{55}&\ding{55}\\

        \midrule
        Significant Accuracy Loss & -- &\checkmark&\ding{55}&\ding{55}&\ding{55}&\ding{55}&\ding{55}&\ding{55}&\ding{55}&\ding{55}&\ding{55}&\ding{55}&\ding{55}&\ding{55}&\ding{55}&\ding{55}&\ding{55}&\ding{55}&\ding{55}&\ding{55}&\ding{55}&\ding{55}&\ding{55}\\
        
        \midrule
        Updating Frequency& -- &\checkmark&\ding{55}&\checkmark&\ding{55}&\ding{55}&\ding{55}&\ding{55}&\ding{55}&\ding{55}&\ding{55}&\ding{55}&\ding{55}&\ding{55}&\ding{55}&\ding{55}&\ding{55}&\ding{55}&\ding{55}&\ding{55}&\ding{55}&\ding{55}&\ding{55}\\
        \bottomrule
    \end{tabular}
    }
\end{table*}

The remainder of this survey is organized as follows. Section \ref{sec:background} reviews the preliminaries of concepts such as FL, CL, and FCL. Section \ref{sec:FL} discusses challenges and solutions associated with FL comonents of FCL. Section \ref{sec:CL} elaborates on difficaulties of CL and the potential solutions. Section \ref{sec:FCL} explains problems that are specific to FCL and overviews the key approaches for mitigating them. Section \ref{sec:future} elaborates on future research directions that are crucial in FCL. Finally, the paper is concluded in Section \ref{sec:conclusion}.

\section{Federated Continual Learning}
\label{sec:background}
The data-driven world requires learning from continuous data in a distributed manner, while respecting privacy. FCL  emerges as a robust solution for scenarios involving both collaborative learning and streaming data. FCL seamlessly integrates FL principles with the capability to manage continuous data streams, offering a versatile and efficient approach to contemporary data challenges. By leveraging FCL, it is possible to achieve the dual objectives of collaboratively training models without centralized data collection and maintaining up-to-date models through CL from streaming data.

The first step toward understanding FCL is to understand its base ideas: FL and CL. FL refers to a variety of model training that involves multiple devices/nodes in collaboration, not sharing raw data with each other for privacy and scalability reasons. On the other hand, CL deals with models' capabilities of incremental learning on non-IID evolving data streams, while retaining knowledge of previous tasks. These two paradigms, while different, meet at the intersection of FCL in the quest to build models that continuously learn from new data without violating the privacy and heterogeneity of decentralized systems. In-depth knowledge of FL and CL puts one in a better perspective to understand the challenges and opportunities that come with FCL.

\subsection{Federated Learning}
FL is a collaborative method for training machine learning models across multiple devices or nodes with the guarantee that raw data would remain on the local devices. It is motivated by the urgent need for privacy-preserving and efficient learning systems in recent distributed environments, such as smartphones, IoT devices, and edge computing networks. In particular, unlike traditional, top-down approaches to learning in a decentralized system, FL lets owners of data participate in training the model without necessarily revealing personally identifiable or commercially sensitive information.

By keeping data decentralized, FL protects user privacy and also allows compliance with strict data protection regulations, such as GDPR. It reduces the risk of data breaches and ensures that the training data remains local; hence, trust among participants can be fostered. FL also leverages the heavy computational power of distributed devices, hence enhancing scalability, and, thus, enabling the creation of models that can make the most out of diverse data sources. This generally results in more generalizable and robust models for applications in which data across users may vary substantially, such as personalized recommendation systems, healthcare, and autonomous systems. FL will be capable of harnessing decentralized data and resources, thus becoming one of the cornerstones for developing intelligent, privacy-aware systems in modern, connected environments.

\subsection{Continual Learning}
In dynamic environments such as real-time sensor networks or streaming applications, data is continuously generated and needs to be processed in a timely manner. For instance, consider a smart city scenario where traffic cameras, weather sensors, and social media feeds constantly produce data streams. In such an environment,which is known as non stationary environment, it is essential for models to be updated continuously with new data to maintain their performance. However, traditional models often struggle to adapt to changes in data distributions, leading to a decrease in performance over time. This challenge is addressed by CL, which focuses on the ability to incrementally learn from new data, while retaining knowledge from previous tasks \cite{10444954, DBLP:journals/corr/abs-1904-07734}. By effectively managing the evolving nature of data, CL ensures that models remain robust and adaptable, preventing the degradation of performance that can occur with static training methods.

CL provides several significant advantages that enhance the effectiveness and adaptability of machine learning models. These key benefits include incremental learning, knowledge transfer, avoiding catastrophic forgetting, adaptability and flexibility, resource efficiency, improved generalization, and application versatility. Firstly, CL allows models to learn incrementally, building on previously acquired knowledge and enabling adaptation to new tasks without forgetting old ones, which is particularly useful in dynamic environments, where data evolves over time \cite{9349197}. Secondly, it facilitates the transfer of knowledge from one task to another, improving learning efficiency for new tasks by leveraging previously acquired skills, thereby leading to faster learning and better performance on related tasks \cite{10.1007/978-3-030-58621-8_23, BOUWMANS20198}.

Furthermore, a main advantage is its ability to mitigate catastrophic forgetting, using techniques such as experience replay, regularization, and dynamic architecture growth to maintain old knowledge \cite{9878681}. Additionally, CL models are highly adaptable, making them suitable for non-stationary, continuously changing real-world applications \cite{HADSELL20201028}. Moreover, they emphasize resource efficiency by reusing existing knowledge and minimizing the need for retraining, which leads to significant savings in computational resources and time. For instance, \cite{NEURIPS2022_4522de41} introduces a method for performing text classification using pre-trained transformers on a sequence of classification tasks, achieving significant memory efficiency and faster inference time compared to state-of-the-art methods, while maintaining competitive predictive performance without requiring extensive retraining. By continually learning from new data, CL models also improve their generalization capabilities, making them more robust to data variations and better at handling unseen scenarios \cite{10365578}. Finally, CL has broad applications across various domains, including robotics, healthcare, and autonomous systems, where the ability to learn and adapt continuously is critical \cite{9289549}.

\section{Challenges of Federated Learning}
\label{sec:FL}
Part of the challenges in FCL are assocciated to the FL structure of this approach as detailed in this section.

\subsection{Heterogeneity}
In collaborative learning, different clients contribute to training, which can enhance overall training performance by leveraging server computational power and improving convergence rates and globalization \cite{10001832,9751161}. However, this diversity can introduce heterogeneity, stemming from factors such as geographical differences, variations in data source distribution, differing data formats, inconsistent data quality, variations in data collection methods, discrepancies in labeling practices, and differences in feature representation \cite{10118862}. Heterogeneity, which refers to the presence of differences and diversity within a set of elements, is one of the prominent challenges in collaborative learning. \cite{10061708} highlights the severe impacts of device and behavioral heterogeneity on model performance, showing substantial degradation in model quality and fairness. This subsection explains different types of heterogeneity and proposes methods to address them effectively, categorizing heterogeneity into two primary types: data heterogeneity and system heterogeneity.

Data heterogeneity arises from the varied nature of data contributed by different clients. Figure \ref{fig:data-heterogeneity} delineates the subsets of data heterogeneity. Table \ref{tab:evaluation} highlights datasets demonstrating such heterogeneity, including non-IID distributions and imbalanced data. Understanding data heterogeneity is crucial for effective model training and integration \cite{Liu2023Predictive, 10174745}. Data heterogeneity is generally classified into statistical and conceptual subsets, each of which presents a distinct challenge that needs to be addressed in detail.

\begin{figure}[t]
    \begin{minipage}{0.3\textwidth} 
    \raggedright 
    \begin{forest} 
    for tree={
        grow'=0,
        parent anchor=east,
        child anchor=west,
        anchor=west,
        inner xsep=0.5mm, 
        align=center,
        l sep=2mm, 
        s sep=2mm, 
        edge path={\noexpand\path[\forestoption{edge}, thick](!u.parent anchor) -- +(5pt,0pt) |- (.child anchor)\forestoption{edge label};},
        font=\footnotesize, 
    }
    [Data\\Heterogeneity
        [Statistical \\Heterogeneity
            [Non-IID Data
                [Changes in \\Input Space]
                [Changes in \\Behaviour]
                ]
            [Imbalanced Data
                [Unequal \\Sample \\Sizes]
                [Skewed Data \\Distribution]
            ]
        ]
        [Conceptual\\ Heterogeneity
            [Label Distribution \\Heterogeneity
                [Diverse \\Label Sets]
                [Inconsistent \\Labeling]
            ]
            [Feature Distribution \\Heterogeneity
                [Varying \\Feature Sets]
                [Feature \\Scaling\\Differences]
            ]
        ]
    ]
    \end{forest}
    \end{minipage}
    \caption{Sub-categories of data heterogeneity in collaborative learning.}
    \label{fig:data-heterogeneity}
\end{figure}

\subsubsection{Statistical Heterogeneity}
Statistical heterogeneity refers to differences in the statistical properties of datasets used by different nodes in a distributed learning system. This can include variations in data distributions, sample sizes, and data quality across different sources. Two critical subsets of statistical heterogeneity are Non-IID (Independent and Identically Distributed) data and imbalanced data.

\paragraph{Non-IID Data}
Many machine learning algorithms assume that the training data are independent and identically distributed (IID), meaning each data point is independent of others and follows the same probability distribution. When data from different clients violate this assumption, they are considered non-IID. These dependencies arise from differences in input space or behavioral differences among clients. 

\begin{table*}[t!]
\centering 
\footnotesize
\caption{Examples of data heterogeneity considered in FCL research. Data Format: categorical (C), numerical (N), and mixed (M). Research field: computer vision (CV), healthcare (Health), cybersecurity (Cyber), Internet of Things (IoT), natural language processing (NLP), and entertainment (EN). Task type: instance incremental (Inst), domain incremental (Dom), task incremental (Task), feature incremental (Feat), and class incremental (Cls). Output type: label (Label), score (S), and partition (Partition). \ding{55} and * denote not applicable and not mentioned.}  
    \scriptsize
    \begin{tabular}{lcccccc} 
    \toprule
    
    Approaches & Data Format & Distribution&Drift Type & Task Type  
    &\makecell{Domain}&Type\\

    \midrule
    
     Fed-ReMECS \cite{NANDI2022340}
    &C&Non-IID&\textasteriskcentered & Inst
    &CV&Label\\
    
    ISCN \cite{10294732}
    &N&Non-IID&\textasteriskcentered & Inst/Feat
    &Cyber&Label\\
    
    \cite{app121910025}
    &C&Non-IID&\textasteriskcentered & Cls/Task
    &CV&Label\\

    FedSpace \cite{10208460}
    &C&Non-IID&R&Task
    &CV&Label\\

   FedConD \cite{9671924}
   &C&Non-IID&R&Dom
   &CV&Label\\

   ASO-Fed \cite{9378161}
   &C&Non-IID&\textasteriskcentered & Inst/Dom&
   CV&Label\\

   \cite{9562751}&
   M&Non-IID&V&Inst&
   EN/NLP&Label\\

   CHFL \cite{9892815}&
   N/M&Non-IID&\textasteriskcentered & Dom/Feat &
   CV/IoT&Label\\

   Cross-FCL \cite{9960821}&
   C&Non-IID&\textasteriskcentered & Task&
   CV&Label\\

   FedDrift-Eager/FedDrift \cite{pmlr-v206-jothimurugesan23a}&
   N&Non-IID&H&Dom&
   \ding{55}&Label\\

   Master-FL \cite{10198306}&
   N/M&Non-IID&R/V&Dom&
   CV&Label\\

   SOFed \cite{10128673}&
   N&Non-IID&V&Inst/Dom&
   CV&Partition\\

   FLARE \cite{10182870}&
   N&Non-IID&H&Dom&
   IoT/CV&Label\\
   
   \cite{NEURIPS2023_d160ea01}&
   N&Non-IID&R&Task/Cls&
   CV&Label\\

    FairFedDrift \cite{salazar2024unveilinggroupspecificdistributedconcept}&
    N&\textasteriskcentered &R&Dom&
    CV&Label\\

    \cite{hu2024energyefficientfederatededgelearning}&
    N&IID/Non-IID&R/V&Inst&
    CV&Label\\
    
    ADMM-FedMeta \cite{10.1145/3466772.3467038}&
    N&Non-IID&H&Task/Cls&
    CV&Label\\
    
    FedINC \cite{10.1145/3625687.3625800}&
    N&Non-IID&H&Inst/Task/Cls&
    CV&Label\\

    Adaptive-FedAVG \cite{9533710}&
    N&IID&H&Task/Cls&
    CV&Label\\

    AFAFed \cite{BACCARELLI2022376}&
    N&IID+Non-IID&R&Inst/Dom/Task&
    IoT&Label\\

    O-GFML/PSO-GFML \cite{9751160}&
    N&Non-IID&\textasteriskcentered &Task&
    \ding{55}&Score\\

    GLFC \cite{Dong_2022_CVPR}&
    N&Non-IID&R&Task/Cls&
    CV&Label\\

    \cite{10193322}&
    N&IID/Non-IID&R&Inst&
    CV&Label\\

    CoOptFL/DYNAMITE \cite{10330725}&
    N&IID/Non-IID&R/V/H&Dom/Task/Inst&
    CV&Label\\

    Fed-IW\&DS \cite{electronics11223668}&
    N&Non-IID&R/V/H&Dom&
    CV&Label\\

    \cite{electronics12040894}&
    M&Non-IID&H&Dom&
    Cyber&Label\\

    FedSKF \cite{electronics13091772}&
    N&IID/Non-IID&R&Cls&
    CV&Label\\

    \cite{10470505}&
    N&IID&R&Dom&
    CV&Label\\

    \cite{10333463}&
    N&Non-IID&\textasteriskcentered &Inst&
    CV&Label\\

    FCIL-MSN \cite{10540639}&
    N&Non-IID&H&Task/Cls&
    CV&Label\\

    FedADC \cite{9517850}&
    N&Non-IID&H&Dom&
    CV&Label\\

    FCL4SR \cite{10309661}&
    M&\textasteriskcentered &H&Dom&
    CV/IoT&Label\\

    \cite{9821057}&
    N&\textasteriskcentered &\textasteriskcentered&Task/Cls&
    Health/IoT&Label\\

   FCL-SBLS \cite{10143925}&
   M&\textasteriskcentered&\textasteriskcentered&Task&
   Cyber&Score\\

   FedStream \cite{10175385}&
   C/N/M&IID/Non-IID&V&\textasteriskcentered&
   CV/Cyber/IoT& Label\\

   FedStream \cite{10198520}&
   N/M& \textasteriskcentered& R/V/H& Inst/Dom&
   Cyber/IoT&Label\\

   \cite{10406164}&
   N&Non-IID&V/H&Dom/Task/Cls&
   CV&Label\\

   FedProK \cite{Gao_2024_CVPR}&
   N&IID/Non-IID&R/H&Task/Feat/Cls&
   CV&Label\\

   FedKNOW \cite{10184531}&N&Non-IID&R/V/H&Task&
   CV&Label\\

   FedRCIL \cite{Psaltis_2023_ICCV}&N&IID/Non-IID&R/V/H&Task&	
   CV	&Label\\
   LCFIL \cite{9973580}	&N&IID/Non-IID&R/V/H&Task/Cls&
   CV&	Label\\

   Re-Fed \cite{Li_2024_CVPR}&N&IID/Non-IID&R/V/H&Dom/Task/Cls&
   CV&Label\\

   ICMFed \cite{math11081867}&N&Non-IID&V&Task&
   CV&Label\\

   LGA \cite{10323204}&N&Non-IID&H&Task/Cls&
   CV&Label\\

   \cite{10295979}&N&IID/Non-IID&V&Inst&
   CV/IoT&Label\\

  Flash \cite{pmlr-v202-panchal23a}&N&Non-IID&R/H&Task&
  CV/NLP&Label\\

  \cite{10097140}&
  M&\textasteriskcentered&\textasteriskcentered&Task&
  NLP	&Label\\

  FedPC \cite{Yuan_2023_CVPR}&
  N&Non-IID&V&\textasteriskcentered
  &CV&Label\\

  cTD-$\alpha$MAML\cite{10148063}&
  N&IID/Non-IID&R/V/H&Task&
  CV/Health&Label\\

  SFLEDS \cite{MAWULI2023119235}&
  N/M&\textasteriskcentered&R/V/H&Inst/Task/Cls&
  Cyber/IoT&Label\\
  
  FedNN \cite{KANG2024110230}
  &N&Non-IID&R/H&Dom&
  CV&Label\\

  Fed-SOINN \cite{ZHU2022168}&
  M&IID/Non-IID&\textasteriskcentered&Task/Cls&
  Cyber&Label\\
  
  RRA-FFSCIL \cite{JIANG2024127956}&
  N&Non-IID&H&Task/Cls&
  CV&Label\\

  \cite{9950044}&
  N&Non-IID&R&Inst&
  CV&Label\\
  
  \cite{YANG202416}&
  N/M&Non-IID&\textasteriskcentered&Task
  &CV/IoT&Label\\

  \cite{LI2024111491}&
  N&IID&H&Dom/Task&
  \ding{55}&Label\\

  FL-IIDS \cite{JIN202457}
  &M&IID&R&Task/Cls&
  Cyber&Label\\

 FedViT \cite{ZUO20241}&
 N&Non-IID&V/H&Task&
 CV&Label\\

 CFeD \cite{ijcai2022p303}
 &N&IID/Non-IID&R/V/H&Dom/Task/Cls&
 CV/NLP&Label\\

 FedET \cite{liu2023fedetcommunicationefficientfederatedclassincremental}
 &N&Non-IID&V/H&Task/Cls&
 CV/NLP&Label\\

 FedWeIT \cite{pmlr-v139-yoon21b}&N&Non-IID&V&Task&
 CV&Label\\
 
 \cite{good2023coordinatedreplaysampleselection}&
 N&\textasteriskcentered&\textasteriskcentered&\textasteriskcentered&
 NLP&Label\\
 
 \cite{hendryx2021federatedreconnaissanceefficientdistributed}&
 N&Non-IID&R/H&Cls&
 CV&Label\\

\cite{10543076}&
N&\textasteriskcentered&\textasteriskcentered&Task&
IoT&Label\\

SLMFed \cite{10399971}&
N&Non-IID&\textasteriskcentered&\textasteriskcentered&
CV&Label\\		

TARGET \cite{10376970}&
N&IID/Non-IID&R/V/H&Task&
CV&Label\\

HFIN \cite{10546981}&
M&Non-IID&R/V/H&Task/Cls
&Cyber&Score\\

 \cite{CHAVES2024101036}&
 N&Non-IID&\textasteriskcentered&\textasteriskcentered&
 CV&Label\\
\bottomrule
    \end{tabular}
\label{tab:evaluation}
\end{table*}

Differences in input space occur when client \(i\) and client \(j\) contribute to the training and their local datasets have different distributions \( P(x_{i}) \neq P(x_{j})\). On the other hand, behavioral differences refer to variations in the way different clients generate, collect, or label their data, leading to different perspectives on events for the machine learning models \( P(y_{i}|x_{i}) \neq P(y_{j}|x_{j})\). These differences can impact the performance of machine learning models trained on such datasets. To address non-IID data as a prominent challenge in distributed systems, the proposed methods are reviewed here to clarify the importance of each approach.

To address changes in the input space through clients, three different methods are proposed. 
Firstly, domain transformation involves techniques designed to align the data distributions from multiple clients to create a more unified input space for a machine learning model. One approach focuses on domains with particular features, leveraging the unique characteristics of each domain to enhance performance. For instance, studies \cite{noauthor_zhao_nodate, 8476540, hoffman_discovering_2012} address the issue that each domain may have its own unique set of features to characterize samples, leading to incompatibilities across domains, and they develop methods to extract a common feature representation. To talk in detail, deep learning models are utilized for learning complex relations between features \cite{8963871}, and domain generalization \cite{8995481}. Moreover, \cite{chen_domain_2021} proposed an approach in which feature transformation and KL divergence minimization are used to align source and target data distributions. Another approach, performed in \cite{siyahjani_supervised_2015, wang_metric_2010, noauthor_weinberger_nodate, JING202039, 8968742}, is domain factorization, which decomposes data into shared and domain-specific components using methods such as matrix factorization. This isolates common patterns and unique features, allowing models to better understand the underlying structure. These techniques improve the robustness and effectiveness of machine learning models across diverse datasets.  
 
Moreover, due to the significant variations in data distributions across different clients, personalization in FL is essential for addressing changes in the input space. This approach involves tailoring a global machine learning model to better fit the unique data and usage patterns of individual users. By doing so, the model becomes not only broadly effective but also finely tuned to the specific needs and behaviors of each user. \cite{9210355} underscores the importance of personalization, while \cite{9743558} explores methods to enhance the performance of FL models by incorporating personalization techniques, thereby addressing the challenges posed by data heterogeneity across different clients. 
Personalization methods are categorized into two subsets of client level personalization and group level personalization. Client level personalization tailors the model to the specific data characteristics and patterns of each individual client.
For instance, \cite{10130784} address the heterogeneity of data across different clients by learning personalized self-attention layers for each client, while aggregating other parameters among clients. Also, \cite{JING2023354}  aims to customize the global machine learning model to better fit the unique data distributions of individual clients. Another idea, explored in \cite{10.5555/3305381.3305498, fallah2020personalized, jiang2023improving}, involves adapting algorithms from the Model-Agnostic Meta-Learning (MAML) framework for use in FL settings. Instead of personalizing the model for each individual client, in group level personalization the model is tailored for groups with homogeneous data distributions. \cite{10081485} leverages group-level personalization by clustering clients based on inference similarity, addressing the non-IID data distribution problem and enhancing the overall performance and efficiency of FL models.

Another approach for addressing the changes in clients' input space is domain adaptation, which addresses the challenge of adapting a model trained on one domain (source domain) to perform well on a different but related domain (target domain) \cite{10017290}. \cite{SUN201584} provides a comprehensive overview of the methods and techniques used in multi-source domain adaptation (MSDA), addressing challenges such as class distribution alignment and domain shifts. There are three different approaches for domain adaptation, dissimilarity methods, sample reweighting, and Generative Adversarial Networks (GANs) \cite{Deng2018,Wang2023,8099799}. Dissimilarity methods aim to reduce the difference between the source and target domain distributions by explicitly measuring and minimizing discrepancies. These methods often focus on aligning the statistical properties of the two domains. For instance, \cite{9430774} proposes the Manifold Maximum Mean Discrepancy (M3D) to measure and minimize local distribution discrepancies. On the other hand, \cite{8578933} introduces a novel method for unsupervised domain adaptation (UDA) that leverages similarity learning to improve the classification performance on an unlabeled target domain using features learned from a labeled source domain. Sample Reweighting is another approach, which involves adjusting the weights of source domain samples to make the source distribution more similar to the target distribution. In particular, \cite{10130085}  proposes a reweighting method for model aggregation in FL that considers the volume and variance of local datasets among clients to address the issue of class label heterogeneity. Also, \cite{10.5555/1613715.1613801,dredze_multi-domain_2010} propose methods to learn across various domains in Natural Language Processing (NLP) tasks by leveraging confidence-weighted parameter combinations, parameter combinations from multiple classifiers. Besides, GANs are used in domain adaptation to generate synthetic data that bridges the gap between the source and target domains. By training a generator and discriminator\cite{8099799}, GANs can create data that is indistinguishable from the target domain, facilitating better adaptation. Accordingly, several papers have contributed significant insights and methodologies, for instance, by leveraging multiple adversarial networks \cite{10.5555/3504035.3504517}, considering a more realistic setting compared to traditional domain adaptation, where it is often assumed that the label spaces of the source and target domains are fully shared \cite{Cao_2018_ECCV}, using deep models, particularly when only a few labeled samples are available in the target domain \cite{NIPS2017_21c5bba1}, and addressing challenges of improving generalization in FL environments, where domain shifts occur \cite{peng2019federated}.
After exploring changes in the input space and the methods for addressing these challenges, it is time to address the changes in the behavior of clients' datasets, which are categorized into four classes of group level personalization methods, contextual information methods, federated multi-task learning, and cohort-based FL \cite{CRIADO2022263}. Group-level personalization effectively addresses the challenges of data heterogeneity and behavioral changes in FL, which is clarified in \cite{9954890, 10.1145/3558005}. By clustering clients based on data similarities and creating specialized models for each group, this approach ensures that models are both accurate and adaptive to changes in client behavior. For instance, \cite{9207469} utilizes hierarchical clustering for local updates in FL to separate clients by the similarity of their updates, which allows for more efficient training on non-IID data and achieving higher accuracy. Moreover, contextual information methods involve techniques such as abnormal client behavior detection \cite{li2019abnormal}, regulated client participation \cite{malinovsky2023federated}, and multicriteria client selection (FedMCCS) \cite{9212434} to handle variations in client behavior and data distribution. These methods enhance the robustness, efficiency, and accuracy of FL by effectively managing non-IID data, client participation variability, and handling resource constraints. To explore more solutions for changes in behavior, Federated Multi-Task Learning (FMTL) has been proposed as an advanced approach within FL that aims at handling client-specific tasks, while maintaining data privacy. This approach is introduced in \cite{NIPS2017_6211080f} by proposing a novel systems-aware optimization method, named MOCHA. It extends the traditional FL by allowing the training of multiple tasks simultaneously across different clients. It offers a robust solution for managing diverse client behaviors by providing personalized, efficient, and fair learning frameworks \cite{9492755}. These approaches improve model accuracy and generalization, while preserving client data privacy \cite{9174890, 9763764}. Furthermore, Cohort-Based FL (CBFL) is a specialized approach within FL that aims to enhance learning performance by grouping clients with similar data distributions into cohorts. In particular, in industrial settings, CBFL can manage skewed data by organizing clients into cohorts based on similar data distributions \cite{HIESSL202264}. This approach improves model performance by aligning the training process with the specific characteristics of industrial clients. \cite{SIKANDAR2023518} proposes a method that is the integration of cohort-based Kernel Principal Component Analysis (KPCA) with multi-path service routing in FL by leveraging KPCA for effective feature extraction and dimensionality reduction, forming client cohorts for homogeneous data distribution, and optimizing communication through multi-path routing, which is a powerful approach to enhance the efficiency and performance of FL systems and addresses the challenges of data heterogeneity and communication efficiency in FL.

\paragraph{Imbalanced Data}
Besides non-IID data, imbalanced data is another subset of statistical heterogeneity, which refers to a situation where certain classes or labels in the dataset are underrepresented compared to others. This imbalance can lead to biased model performance, favoring the majority class and neglecting the minority class. Imbalanced data is categorized into two subsets, which are unequal sample sizes and skewed data distribution. When different classes in the dataset have a vastly different number of samples causes unequal sample sizes, leading to fail in correctly classifying the minority class due to the overwhelming presence of the majority class. This can result in high accuracy for the majority class but poor performance for the minority class, which is often more critical in applications such as fraud detection, intrusion detection, or disease diagnosis. To address this challenge, techniques such as under-sampling the majority class or over-sampling the minority class are commonly used, which is explored in \cite{9078901, 4717268, 10.1007/978-3-031-43415-0_22}. On the other hand, skewed data distribution refers to the asymmetry in the data distribution for a particular variable or feature, which means that the frequency of occurrence of some classes is significantly higher than others, leading to a disproportionate representation of classes, making it difficult for models to learn the decision boundary for the minority class. This often results in poor generalization of new data, especially for the minority class. Addressing skewed distributions involves strategies such as clustering and density-based approaches to balance the data before training \cite{MIRZAEI2021114035}. The upper part of Figure \ref{fig:data-heterogeneity}, statistical heterogeneity, is addressed through this part, in which most of the solutions are the subsets of FL. In the following part the conceptual heterogeneity and its subsets are addressed.

\subsubsection{Conceptual Heterogeneity}
Conceptual heterogeneity arises from differences in data distribution, computational power, and network conditions among participating nodes, leading to challenges in achieving efficient and robust training performance. This discussion focuses on the subset of conceptual heterogeneity related to data heterogeneity, where clients have different types of data. These differences manifest in label distribution or feature distribution, as depicted in the middle part of Figure \ref{fig:data-heterogeneity}. When considering label distribution heterogeneity, different clients may have datasets with varying sets of labels (diverse label sets). For example, one node might have data labeled for dog breeds, while another has labels for car models, complicating the training of a unified model. To address this, \cite{NIPS2017_6211080f} proposes FL along with a multi-task learning approach to enable each client to learn its specific task,  while simultaneously sharing and benefiting from a common representation or model, thereby improving both individual and global learning outcomes. Besides, \cite{pmlr-v37-ganin15} by proposing an unsupervised domain adaptation approach aims to improve the performance of a model on a target domain (which has no labeled data) by leveraging labeled data from a source domain. Finally, \cite{5288526} reviews transfer learning, in which the core idea is to transfer the representations, parameters, or knowledge from the source task to the target task, thereby enhancing the model's performance on the target task despite differences in label distributions.

This helps in dealing with diverse label sets by accommodating varying tasks and labels. Additionally, inconsistencies in how data is labeled across different clients (inconsistent labeling) can occur. For instance, one node might label an image as "cat", while another node labels a similar image as "feline", leading to confusion during model training. \cite{pmlr-v54-mcmahan17a} introduces an approach in which clients can periodically align their labeling schemes through consensus mechanisms, ensuring that the labels become more consistent over time. This can be done through FL protocols that involve occasional exchanges of labeled examples or metadata. Also, \cite{pmlr-v37-long15} addresses inconsistent labeling by aligning feature distributions between source and target domains using deep adaptation layers and domain adversarial training (label normalization and standardization). This creates a shared feature space, minimizing the impact of label discrepancies. Semi-supervised learning is another approach, which is reviewed in \cite{van_engelen_survey_2020}, to address inconsistent labeling by leveraging a small amount of labeled data along with a large amount of unlabeled data to improve the overall labeling quality. Feature distribution heterogeneity occurs when clients have different features available in their datasets (varying feature sets) or when the scale of features varies across nodes (feature scaling differences). For example, in a healthcare system using datasets from several hospitals, one hospital might record patient data with features such as age and weight, while another hospital records features such as blood pressure and cholesterol levels, complicating model integration. Methods such as feature alignment techniques \cite{hotelling_relations_1992}, data augmentation \cite{shorten_survey_2019}, and collaborative feature selection \cite{chandrashekar_survey_2014} are utilized for addressing this challenge. Feature scaling differences arise if one hospital records blood pressure in mmHg, while another records it in kPa. In this case, standardization and normalization \cite{824819}, Federated Preprocessing \cite{DBLP:journals/corr/abs-1806-00582}, and Metric Learning \cite{DBLP:journals/corr/BelletHS13} are proposed for addressing this. 

\subsubsection{System Heterogeneity}
System heterogeneity in FL arises from differences in client hardware and software, impacting training efficiency and reliability. Hardware heterogeneity includes variations in computational power, memory, storage, and network connectivity. High-end devices process data efficiently, while resource-constrained nodes may struggle with large models, causing synchronization delays. Network disparities further affect data transmission rates and system performance.

Software heterogeneity stems from diverse operating systems (e.g., Android, iOS, Windows, and Linux), software versions, and programming frameworks (e.g., TensorFlow, PyTorch, and Java). Inconsistent software versions can lead to execution issues, requiring strict version control and interoperability solutions. Addressing these challenges is essential for ensuring seamless operation, consistent performance, and robustness in FL and distributed systems.

\subsection{Resource Constraints}
Collaborative machine learning, where multiple users contribute their computational power and data, is promising but faces significant challenges due to resource constraints. These constraints can limit the efficiency and effectiveness of the collaborative learning process. Table \ref{tab:dataset} lists datasets with structural variations and drift types, relevant for studying resource limitations. Accordingly, \cite{9112742} states that heterogeneity and limited computation and communication resources are the key factors that challenge machine learning at the network edge. These constraints hinder efficiency in the application of edge servers in collaborative learning processes, most especially under edge-cloud collaborative machine learning systems. This makes achieving an optimal balance between learning performance and resource consumption challenging. Additionally, in Federated Edge Learning (FEEL), the high communication costs due to the involvement of numerous remote devices pose a significant resource constraint. Efficient algorithms that speed up convergence can help alleviate these costs, which is addressed by \cite{9609654}. Moreover, collaborative machine learning techniques such as FL often suffer from low resource utilisation due to sequential computation and communication, which limits their efficiency. Leveraging pipeline parallelism can significantly improve resource utilisation and reduce training time \cite{9893091}. Moreover, dynamic resource allocation and task offloading in multi-access edge computing (MEC) environments can help overcome resource constraints by optimizing the use of available resources and minimizing energy consumption, while meeting delay constraints \cite{9916157}.

\subsection{Communication Overhead}
In collaborative machine learning, multiple devices work together to train a model. This process, while beneficial, introduces significant communication overhead, which can hinder efficiency and scalability. Effective strategies to reduce communication overhead are crucial for optimizing the collaborative learning process. Several factors lead to communication overhead, including communication-efficient FL, local updates before communication, compression techniques, selective communication strategies, and decentralized learning algorithms. Frequent model updates in FL  involve continual exchange of updates among devices, leading to significant communication delays. Strategies such as probabilistic device selection, quantization methods, and efficient wireless resource allocation can help reduce to communication load and improve convergence speed \cite{doi:10.1073/pnas.2024789118, 8889996, 10.1007/978-3-031-02462-7_21}. These strategies collectively enhance FL's performance and scalability by mitigating communication overhead.

\subsection{Model Stability}

When multiple clients participate in the training process, it is crucial to consider the global model's ability to maintain consistent and reliable performance as it is updated over multiple rounds with data from various distributed sources. Key aspects include consistent performance (avoiding large fluctuations), convergence (gradual stabilization of parameters), and scalability (handling an increasing number of clients without performance degradation). These factors contribute to a robust model that generalizes properly across diverse client data.

Several techniques enhance model stability in FL, including Federated Averaging (FedAvg), regularization, learning rate scheduling, data augmentation, client selection, and adaptive aggregation. FedAvg \cite{pmlr-v54-mcmahan17a} improves stability by averaging local updates, reducing communication overhead, and handling non-IID data. Regularization techniques (e.g., L2 regularization, and dropout) help to prevent overfitting and support smoother convergence, with dynamic regularization shown to enhance FL robustness \cite{acar2021federated}. Learning rate scheduling ensures steady progress and prevents instability.

Data augmentation enhances training diversity, while preserving privacy. For instance, FedM-UNE \cite{10214273} applies augmentation without transferring raw data, improving model stability. Client selection strategies balance contributions from different data distributions by prioritizing clients with sufficient computational and communication resources \cite{8761315}, optimizing training efficiency. Adaptive aggregation methods, which weight model updates based on data quality or quantity, further improve robustness and scalability \cite{8664630}. These strategies collectively contribute to a more stable and reliable FL framework.

\subsection{Privacy Preservation}
When different clients participate in training models, privacy preservation is essential to protect sensitive data, comply with regulatory requirements, maintain user trust, mitigate risks, and enhance data utility. By implementing robust privacy-preserving techniques, organizations can ensure the secure and responsible handling of data in distributed environments, fostering a safer and more trustworthy digital ecosystem \cite{singh_framework_2022}. Several techniques and frameworks have been developed to address these concerns effectively.

\paragraph{Homomorphic Encryption (HE)}
It is a powerful cryptographic mechanism that enables computations to be performed directly on encrypted data, eliminating the need for decryption during processing. This approach ensures that data privacy is maintained throughout the computational process. In the context of FL, HE can be applied to encrypt local model parameters before they are transmitted to a centralized server. The server then aggregates these encrypted parameters without decrypting them, thereby preserving the privacy of individual data points and preventing any potential exposure of sensitive information \cite{app12020734, fi13040094, fi15090310}.

\paragraph{Differential Privacy (DP)}
It is a technique that enhances data privacy by adding controlled noise to the data or model parameters, thereby preventing the extraction of sensitive information while preserving the overall utility of the data. In FL, Local Differential Privacy (LDP) can be employed to add noise to the data before it leaves the user's device. This method ensures that the central server, which aggregates the data from various users, cannot infer any private information from the aggregated data, effectively maintaining user privacy. For instance, \cite{9069945} introduces a framework where artificial noise is added to client parameters before aggregation to ensure privacy. This method, called "noising before model aggregation FL" (NbAFL), satisfies differential privacy by adjusting noise variance. The authors also develop a theoretical convergence bound, showing a trade-off between privacy and performance, and propose a K-client random scheduling strategy to optimize this trade-off. Evaluations confirm that the theoretical results match simulations, aiding the design of privacy-preserving FL algorithms.

\paragraph{Secure Multi-Party Computation (SMC)}
This approach allows multiple parties to collaboratively compute a function over their inputs, while ensuring that each party's inputs remain private. This mechanism ensures that no single party gains access to all the data, preserving the confidentiality of individual inputs. In FL frameworks, SMC can be employed to securely aggregate model updates from multiple users, thereby preventing the exposure of their individual data. This method provides a robust guarantee of data privacy, making it a vital tool for maintaining security and trust in collaborative machine learning environments \cite{10.1145/3338501.3357371}.

\begin{table*}[t]
\centering
\caption{List of datasets used in studies on FCL. Parameters: trial (T), channel (CH), pixel (P), and features (F). Repository: Queen Mary University of London
(QMUL), Electronic Engineering and Computer Science (EECS), University of Toronto (UofT), National Institute of Standards and Technology (NIST), University of California, Irvine (UCI), Stanford University (SU), Princeton University (PU), University of New South Wales (UNSW), Canadian Institute for Cybersecurity (CIC), Ruhr University Bochum (RUB), University of Minnesota (UMN), California Institute of Technology (Caltech), University of New Brunswick (UNB),
University of Extremadura (UEx), Northwestern Polytechnical University (NWPU), University of Edinburgh (UoE), Stanford University (SU), Xidian University (XDU), University of Minnesota (UMN), University of Amsterdam (UvA), Zalando Research (ZR), and Littelfuse Incorporated (LFI). Drift type: gradual (G) and abrupt (A). Data type: non-tabular (N) and multivariate (M).}
\resizebox{\textwidth}{!}{  
\begin{NiceTabular}{lccccccccccc} \toprule
    \addlinespace[5pt]
    Dataset & Repository&\#Instances&\#Features&\#Classes& Dimensionality&Data Type&Drift Type&Time-series & Data Format&Structure&Descriptions\\
    \addlinespace[5pt]
    \midrule
    \addlinespace[5pt]
    \textbf{Real World Dataset}&&&&&&&&&&&\\
    \addlinespace[5pt]
    
    CIFAR-10 \cite{krizhevsky_learning_nodate} &UofT
    &60K&3072&10 &32P x 32P x 3CH&N&\ding{55}&\ding{55}&Image&S&Image Classification\\
    \addlinespace[2pt]

   CIFAR-100 \cite{krizhevsky_learning_nodate} &UofT
    &60K&3072&100 &32P x 32P x 3CH&N&\ding{55}&\ding{55}&Image&S&Image Classification\\
    \addlinespace[2pt]
    MNIST \cite{726791}&NIST
    &70K&784&10&28P × 28P&N&\ding{55}&\ding{55}&Image&S&Digit Classification\\
    \addlinespace[2pt]
    Fashion-MNIST \cite{xiao2017fashionmnistnovelimagedataset}&ZR
    &70K&784&10&28P × 28P&N&\ding{55}&\ding{55}&Image&S&Clothing  Classification\\
    \addlinespace[2pt]
    EMNIST \cite{7966217}&NIST
    &2M&784&62&28P × 28P&N&\ding{55}&\ding{55}&Image&S&Letters/Digit Classification\\
    \addlinespace[2pt]
    SVHN \cite{netzer_reading_nodate}&SU
    &630,388&3072&10&32P × 32P × 3CH&N&\ding{55}&\ding{55}&Image&S&Digit Recognition\\
    \addlinespace[2pt]
    GTSRB \cite{6033395}&RUB
    &50K&3072&43&32P × 32P × 3CH&N&\ding{55}&\ding{55}&Image&S&Traffic Sign Recognition\\
    \addlinespace[2pt]
    FitRec \cite{10.1145/3308558.3313643}&UMN
    &102/168/253K&-&-&-&M&\ding{55}&\checkmark&Tabular&S&Recommendation\\
    \addlinespace[2pt]
    Adult&UCI
    &48842&14&2&14F&M&\ding{55}&\ding{55}&Tabular&S&Classification\\
    \addlinespace[2pt]
    RCV \cite{lewis_rcv1_nodate}&Reuters&804414&47236&	103	&-&N&\ding{55}&\ding{55}&Text&S&News Classification\\
    
    \addlinespace[2pt]
  
    Caltech-256	&Caltech
    &	30,607&*&256&* × * ×3CH&N&\ding{55}&\ding{55}&Image&S&	Object Recognition\\
    \addlinespace[2pt]
  
    ImageNet \cite{5206848}&SU/PU
    & 14M&196608&21K&256P x 256P x 3CH&N&\ding{55}&\ding{55}&Image&S&Visual Object Recognition\\
    \addlinespace[2pt]
    
    TinyImageNet \cite{le_tiny_nodate}&SU/PU
    &120K&12288&200&64P x 64P x 3CH&N&\ding{55}&\ding{55}&Image&S&Visual Object Recognition\\
    \addlinespace[2pt]  

    MiniImageNet&\ding{55}&60K&21168&100&84P × 84P × 3CH&N&\ding{55}&\ding{55}&Image&S&Few-shot Classification\\
    \addlinespace[2pt]  
   
    ToN-IoT \cite{pmlr-v54-mcmahan17a}&UNSW
    &2.3M&44&10&44F&M&\ding{55}&\ding{55}&Tabular&S&Intrusion Detection\\
    
    \addlinespace[2pt]
    CIC-IDS2017 \cite{10.1007/978-3-030-25109-3_9}&CIC
    &3M&80&15&80F&M&\ding{55}&\ding{55}&Tabular&S&Intrusion Detection\\
    \addlinespace[2pt]
    Aloi&UvA
    &110250&82944&1K&192Px144Px3CH&N&\ding{55}&\ding{55}&Image&S&Object Detection\\

    \addlinespace[2pt]
    Helena&UCI
    &65196&28&2&28F&M&\ding{55}&\ding{55}&Tabular&S&Binary Classification\\
    \addlinespace[2pt]   
    Jannis &UCI
    &83733&54&4&54F&M&\ding{55}&\ding{55}&Tabular&S&Performance Prediction\\
    \addlinespace[2pt]
    NWPU-RESIS45&NWPU
    &31500&	196,608&	45&	256P×256P×3CH	&M&\ding{55}&	\ding{55}	&Image&	S	&Sensing Classification\\
    \addlinespace[2pt]   
    CINIC-10 \cite{DBLP:journals/corr/abs-1810-03505}&	UoE
    &270K&	3,072&	10	&32P×32P×3CH&	N&\ding{55}&	\ding{55}&	Image&	S&Image Classification\\
    \addlinespace[2pt]   
    Twitter Sentiment&SU
    &1.6M&6&3&6F&M&\ding{55}&\ding{55}&Text&S&Sentiment Analysis\\
    \addlinespace[2pt]
    MLRSNet \cite{QI2020337}&XDU
    &109161&196,608&46	&256P x 256P x 3CH	&M&\ding{55}&	\ding{55}&	Image&	S&Sensing Classification\\
    \addlinespace[2pt]
    Movielens \cite{10.1145/2827872}&UMN
    &0.1/1/20M&6/8F&5&6/8&M&\ding{55}&\ding{55}&Tabular/Text&S&Movie Ratings\\
    \addlinespace[2pt]    
    Sensorless Drive Diagnosis&LFI/UCI
    &58,509&48&11&48F&N&\ding{55}&\ding{55}&Tabular&S&Motor Fault Diagnosis\\
    \addlinespace[2pt]    
    FaceScrub \cite{7025068}&Exposing.ai
    &105830&128&530&* × * ×3CH&N&\ding{55}&\ding{55}&Image&U&Face recognition\\
    \addlinespace[2pt]
    SocNav1 \cite{data5010007}&UEx
    &1200&	6F&	\ding{55}&	6F&	M&\ding{55}&	\ding{55}&	Tabular&	S&	Social Navigation \\
    \addlinespace[2pt]    
    BoTIoT&UNSW	
    &72M	&46&	4	&46F&	M&\ding{55}	&\ding{55}	&Tabular	&S&	Intrusion Detection\\
    \addlinespace[2pt]    
    NSL-KDD \cite{choudhary_analysis_2020}&UNB&148517&42&5&42F&M&\ding{55}&\ding{55}&Tabular &S&Intrusion Detection\\
    
    \addlinespace[2pt]  
    Forest Covertype \cite{BLACKARD1999131}&UCI
    &581012&54&7&54F&M&\ding{55}&\ding{55}&Tabular&S& Classification\\
    
    \addlinespace[2pt]    
    DEAP \cite{5871728} &QMUL-EECS
    &8064&44&4&40T × 40CH&M&\ding{55}&\checkmark&Tabular&S&Emotion Analysis\\
    \addlinespace[5pt]
    \midrule
    \addlinespace[5pt]
    \textbf{Synthetic Dataset}&&&&&&&&&&&\\
    \addlinespace[5pt]
    SINE \cite{pesaranghader_fast_2016, pesaranghader2018mcdiarmiddriftdetectionmethods, pesaranghader_reservoir_2018}&\ding{55}	&*&2	&2	&2F	&N	&A	&\checkmark	&Tabular&S	&Binary classification\\
    \addlinespace[2pt]
    
    Circles \cite{pesaranghader_fast_2016, pesaranghader2018mcdiarmiddriftdetectionmethods, pesaranghader_reservoir_2018}&\ding{55}&*&2	&2	&2F&N&G&\ding{55}&Tabular&S	&Binary classification\\
    
    \addlinespace[5pt]

     \bottomrule
\end{NiceTabular}
    }
\label{tab:dataset}
\end{table*}

\paragraph{Secure Aggregation}
Secure aggregation protocols ensure that individual updates from local models are securely aggregated without exposing individual contributions. These protocols are designed so that the central server has access only to the aggregated model parameters, not the individual updates from each participant. This approach preserves user privacy by preventing the central server or any other party from accessing or inferring the private data of individual users \cite{DBLP:journals/corr/abs-1901-09888}. Such protocols are crucial in FL frameworks, where maintaining the confidentiality of user data, while still enabling collaborative model training.

\section{Challenges of Continual Learning}
\label{sec:CL}
Despite the advantages of CL, streaming data inherently leads to concept drift, a phenomenon where the underlying data distribution changes over time, which can lead to a decrease in model performance as the data distribution evolves. 

\subsection{Concept Drift}
Concept drift can be categorized into three main types: virtual drift, real drift, and hybrid drift. Table \ref{tab:dataset} provides examples of datasets designed to observe gradual and abrupt concept drifts.

\subsubsection{Virtual Drift}
Virtual drift, also known as temporary drift \cite{lazarescu_using_2004}, sampling shift \cite{Salganicoff1997}, and feature change \cite{doi:10.1137/1.9781611972771.1}, refers to changes in the distribution of input data \( P(X) \) over time without altering the underlying relationship between inputs and outputs. For instance, consider an e-commerce website, where the traffic sources initially consist of 50\% from search engines, 30\% from social media, 10\% from email campaigns, and 10\% from direct visits.

Over time, due to new marketing strategies, this distribution shifts to 30\% from search engines, 50\% from social media, 10\% from email campaigns, and 10\% from direct visits. This shift, while not changing the relationship between traffic and sales, can cause the predictive model based on the original data to perform poorly. To maintain accuracy, the model must be updated to reflect the new distribution of traffic sources.

\subsubsection{Real Drift}

\begin{figure*}[t]
  \centering
  \includegraphics[trim={0.2cm 0.3cm 0.2cm 0.2cm},clip,width=\textwidth]{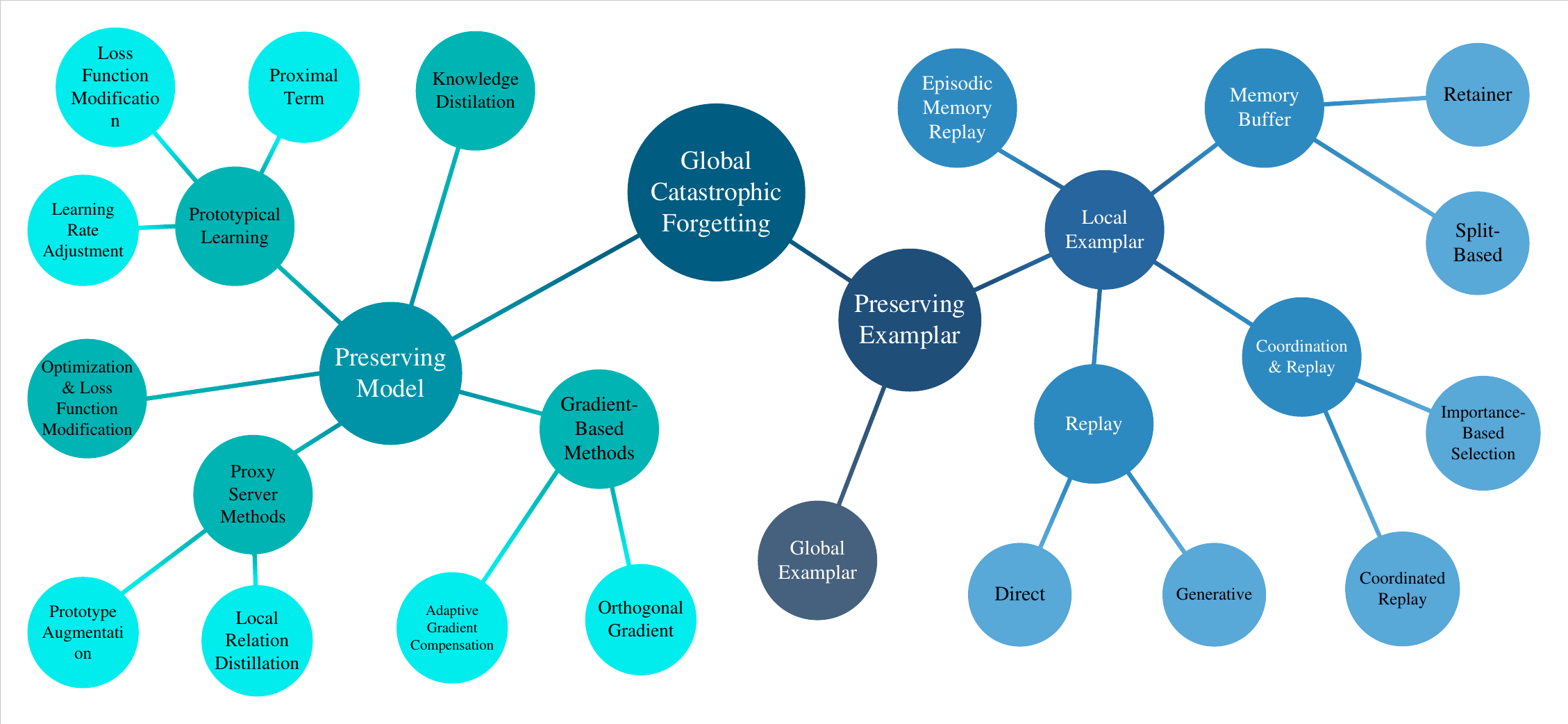}
  \caption{Taxonomy of approaches used to address global forgetting. While this taxonomy is created based on the literature on CL, it can be applied to FCL, as it also uses the principles of CL.}
  \label{fig:forgetting}
\end{figure*}

On the other hand, real drift occurs when the relationship between the input data and the target variable \( P(y|X) \) changes over time. This means the underlying concept that the model is trying to learn is itself changing. For instance, in a financial market prediction model, the factors that influence stock prices may change due to new regulations, economic policies, or shifts in market dynamics. As a result, the relationship between the input features (e.g., economic indicators) and the target variable (e.g., stock prices) changes.

\subsubsection{Hybrid Drift}
Hybrid drift involves both virtual and real concept drifts occurring simultaneously. This means that there are changes in the data distribution \( P(X) \) as well as changes in the underlying relationship between the features and the target variable \( P(y|X) \). Hybrid concept drift is particularly challenging because it requires models to adapt to both types of changes simultaneously. For example, in a recommendation system for an online store, the types of products offered (input distribution) might change, and at the same time, customer preferences (relationship between inputs and outputs) might also change. 

After exploring the various changes in data distribution caused by streaming data, this section proposes and addresses the different challenges that occur in non-stationary environments, including dealing with limited data, managing class incremental learning, and mitigating catastrophic forgetting.

\subsection{Solutions}
To effectively address these changes over time, Figure \ref{fig:forgetting} presents a comprehensive taxonomy of concept drift and outlines solutions for each category. To address virtual concept drift, memory-based methods are proposed to help models retain the knowledge of previously seen data to adapt to changes over time effectively \cite{8496795}. These methods store a subset of past data and periodically retrain the model using this data. These methods can be categorized into: rehearsal methods and generative methods. On the one hand rehearsal methods store a subset of previous data and mix it with new data during training, which can prevent catastrophic forgetting. For instance, \cite{chaudhry2019tiny} explores the use of small episodic memory buffers to mitigate catastrophic forgetting in CL scenarios. Also, a method called Maximal Interfered Retrieval (MIR) is proposed in \cite{NEURIPS2019_15825aee}, which selectively rehearses past experiences that are most likely to suffer interference from new data. Accordingly, \cite{NEURIPS2019_e562cd9c} utilizes gradient information to select training samples with the most significant impact on model parameters. On the other hand, instead of storing actual data, generative methods generate synthetic data that mimics past data distributions, which helps in reducing storage requirements. For instance, \cite{NEURIPS2021_b4e267d8} utilizes meta-learning to enable the model to learn continually. Inspired by human concept learning, a generative classifier is developed that effectively leverages data-driven experiences to learn new concepts from just a few samples, while remaining immune to forgetting. This approach enhances the model's adaptability and robustness in CL environments, ensuring sustained performance and knowledge retention over time. \cite{NIPS2017_0efbe980} proposes an approach, which involves training a generative model to produce samples that mimic previously seen data, which are then used alongside new data to update the main learning model. This technique allows the model to maintain performance on old tasks, while learning new ones, effectively enabling CL without significant performance degradation. It is important that previously learned input domains of data are not forgotten. Regularization methods are proposed to address virtual drift by imposing constraints on the weight updates. For instance, the method in \cite{WANG2022104509} focuses on dynamically adjusting the importance of features based on their relevance to the current data distribution.

To address the changes in the distribution of target data (real drift), contextual information methods and architecture-based methods are proposed. Contextual information methods utilize additional data, often external to the main dataset, to provide a deeper understanding of the context in which the data was generated. This can help in identifying patterns and changes in data distribution over time. \cite{pmlr-v80-serra18a} leverages contextual information by implementing a hard attention mechanism to focus on task-relevant information, thereby improving the model's ability to retain previous knowledge, while learning new tasks without the loss of crucial previously acquired information. 
Additionally, architecture-based methods address changes in the underlying data distribution (real concept drift) by modifying the model's structure to adapt to these changes effectively. For instance, \cite{Mallya_2018_ECCV} adapts a single fixed neural network to multiple tasks by learning binary masks that selectively activate weights, thus allowing the network to maintain high performance on both new and previously learned tasks without suffering from catastrophic forgetting. This method incurs minimal overhead and is effective for various classification tasks, even those with significant domain shifts. \cite{Mallya_2018_CVPR} addresses real concept drift by iterative pruning and retraining weights in a neural network, allowing it to accommodate new tasks, while preserving performance on prior tasks. This method compresses the model, freeing up capacity for new tasks and ensuring CL without performance degradation. In the next subsection, the requirements for datasets to observe different types of drifts are discussed and explored.

\subsection{Catastrophic Forgetting and Continual Learning Dilemmas}
\begin{figure*}[t]
  \centering 
  \includegraphics[width=0.65\textwidth ,clip]{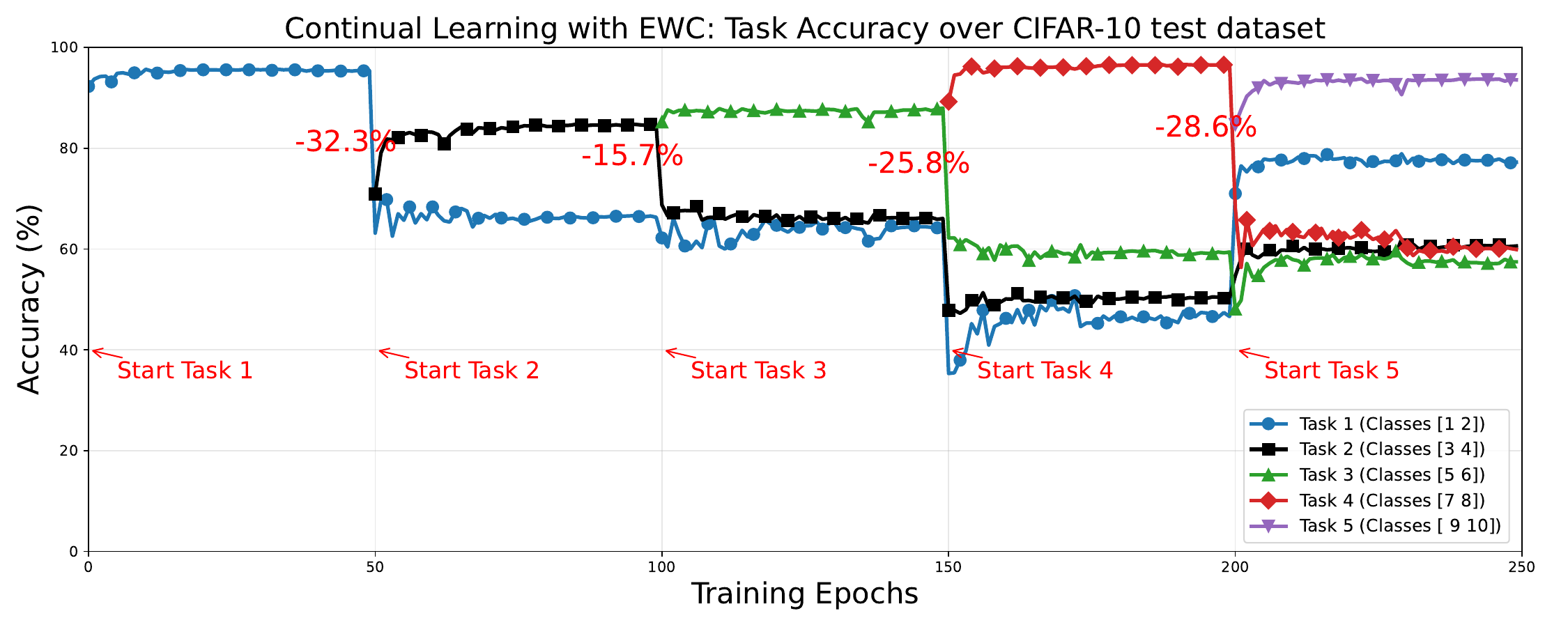}
  \caption{Task-wise accuracy evolution in CL using EWC on CIFAR-10: The plot illustrates the performance trajectory of a CL model trained using EWC on the CIFAR-10 dataset. The dataset is partitioned into five sequential tasks, each comprising two classes. As training progresses through tasks, the accuracy on earlier tasks progressively declines, highlighting the catastrophic forgetting phenomenon. EWC attempts to mitigate this by preserving important parameters from previous tasks, yet the model still experiences notable performance drops, as indicated by the percentage reductions annotated on the plot. This visualization underscores the challenges of maintaining stability in CL settings, even with regularization-based approaches like EWC.} 
  \label{fig:EWCCF}
\end{figure*}

Catastrophic forgetting refers to the tendency of a model to forget previously learned information when acquiring new knowledge. Figure\ref{fig:EWCCF} highlights catastrophic forgetting in CL by visualizing task-wise accuracy evolution in CL using Elastic Weight Consolidation (EWC) algorithm on CIFAR-10 dataset. 

\begin{figure}[t]
  \centering
  \includegraphics[width=\columnwidth ,trim={2.7cm 6cm 1cm 2cm},clip]{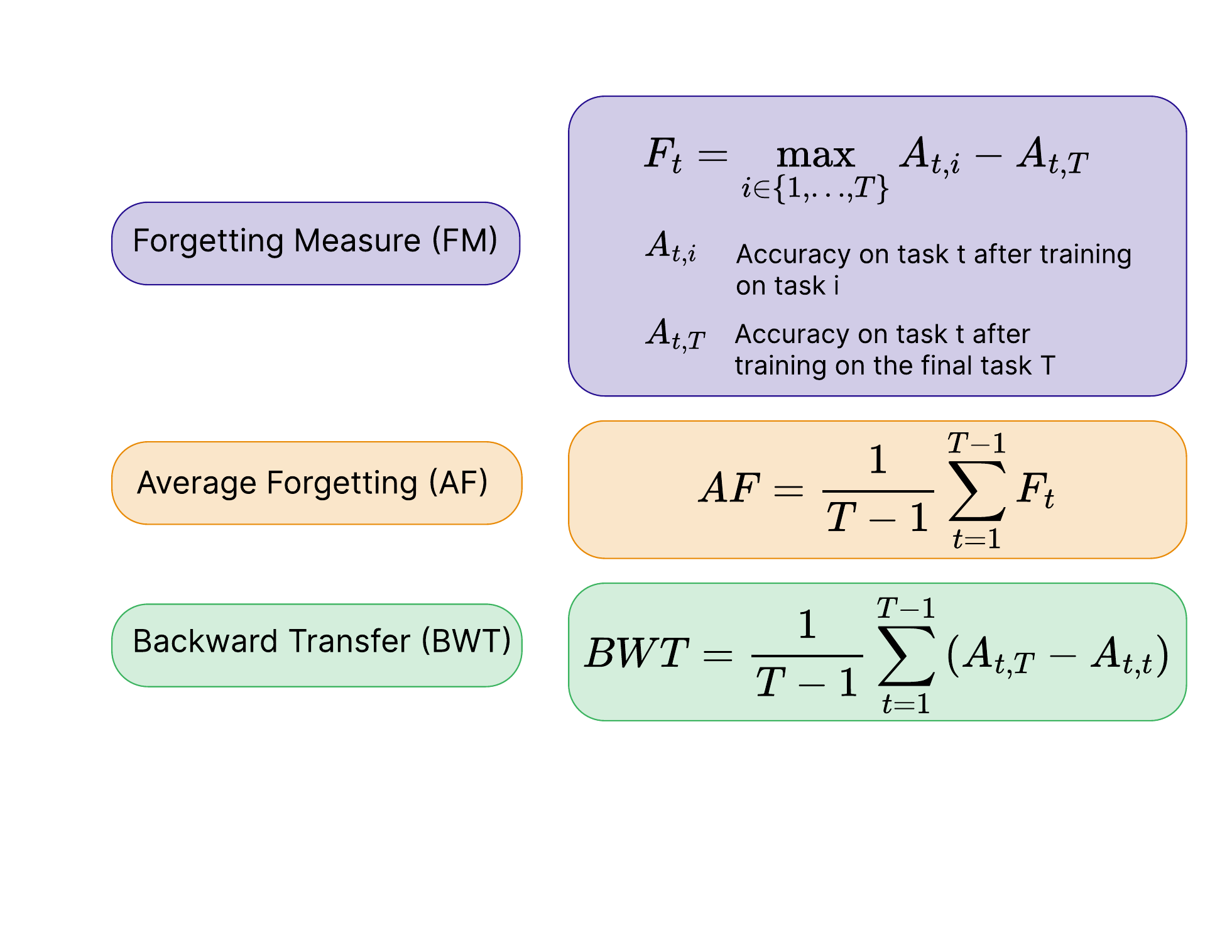}
  \caption{Measures commonly used for catastrophic forgetting}
  \label{fig:CFMeasures}
\end{figure}

To effectively assess catastrophic forgetting, researchers employ several well-established metrics that quantify the extent to which a model loses knowledge of previously learned tasks, when acquiring new ones. Figure~\ref{fig:CFMeasures} illustrates three key metrics commonly used in CL literature.

First, the Forgetting Measure (FM) is computed individually for each task and reflects the drop in performance from the task’s highest accuracy (immediately after training) to its final accuracy (after learning all tasks). A higher FM value indicates a greater degree of forgetting during training~\cite{Chaudhry_2018_ECCV}.

Second, Average Forgetting (AF) aggregates the forgetting measures across all tasks, providing a single scalar value that summarizes the model’s forgetting behavior. This global metric is particularly useful for comparing the effectiveness of different CL methods~\cite{Chaudhry_2018_ECCV}.

Lastly, \underline{B}ack\underline{w}ard \underline{T}ransfer (BWT) captures how learning new tasks influences the performance on previous tasks. BWT can be either positive, indicating knowledge transfer that improves earlier task performance, or negative, signifying catastrophic forgetting~\cite{Chaudhry_2018_ECCV, díazrodríguez2018dontforgetforgettingnew, 10.5555/3295222.3295393}.

The broader implications of catastrophic forgetting are further explored in Section~\ref{sec:FCL}, which categorizes challenges and solutions within the FCL domain. 

\begin{figure}[h!]
  \centering
  \includegraphics[width=\columnwidth ,clip]{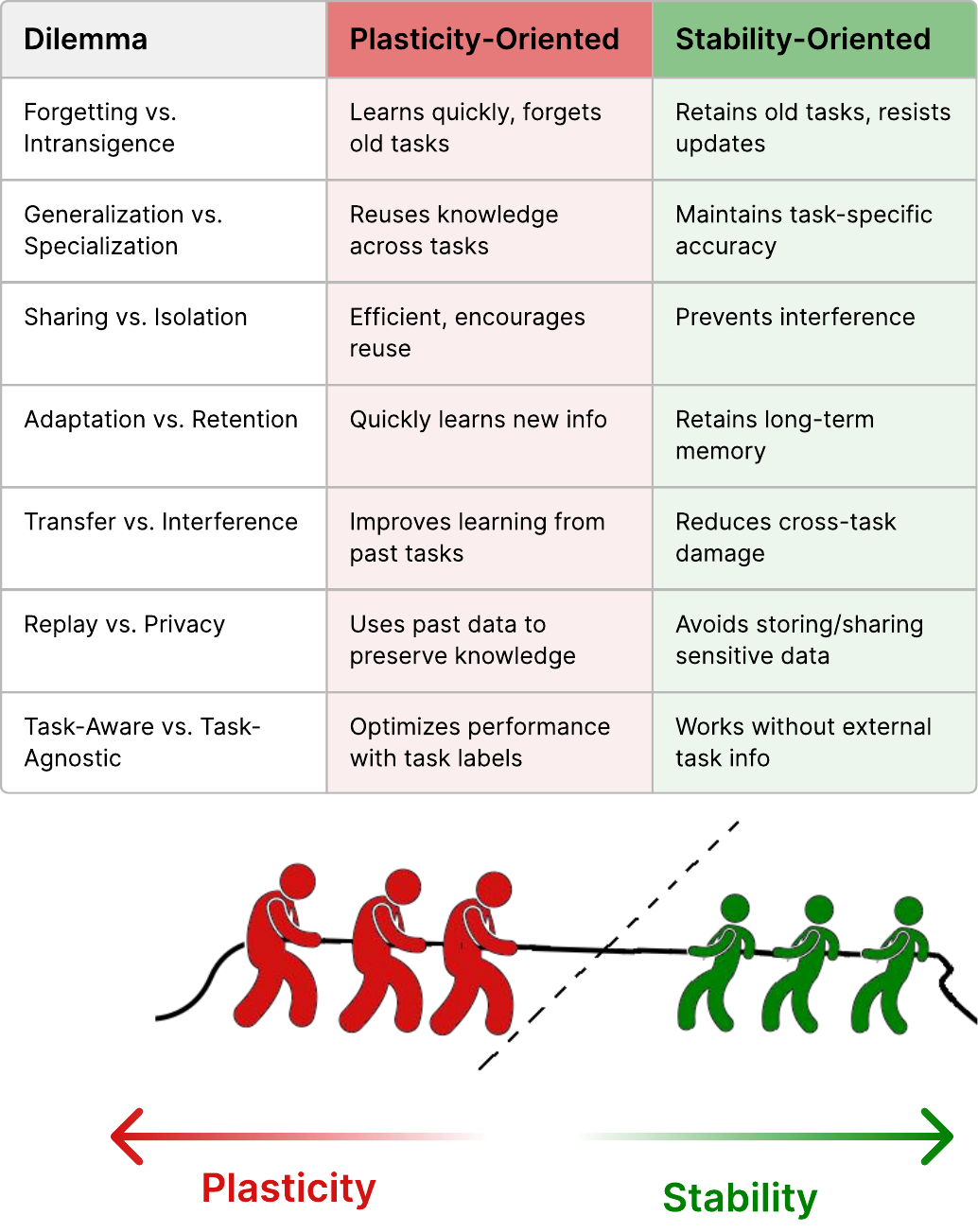}
  \caption{The Stability–Plasticity Trade-off in Continual Learning: A Dilemma-Centric Perspective}
  \label{fig:CLDilemmas}
\end{figure}

In contrast to catastrophic forgetting, which reflects a model’s loss of previously acquired knowledge, intransigence denotes the model’s difficulty in learning new tasks. This typically results from an excessive bias toward earlier tasks, often introduced by strong regularization techniques or heavy reliance on prior task-specific knowledge \cite{Chaudhry_2018_ECCV}. Such constraints promote stability but suppress plasticity, limiting the model’s capacity to adapt to novel information. The coexistence of forgetting and intransigence underscores the fundamental stability–plasticity dilemma in CL \cite{10444954, 9349197, Chaudhry_2018_ECCV, NEURIPS2020_518a38cc, 10203470}. This dilemma encapsulates the inherent trade-off between stability, which safeguards learned knowledge, and plasticity, which facilitates adaptation to new experiences (see figure~\ref{fig:CLDilemmas}).

Figure~\ref{fig:CLDilemmas} also presents a detailed taxonomy of key dilemmas that arise from the stability and plasticity trade-off in CL. Each row in the table contrasts the goals and implications of plasticity-oriented and stability-oriented approaches across various dimensions. For instance, while plasticity favors quick adaptation and knowledge transfer across tasks, stability ensures retention, privacy preservation, and task-specific accuracy. Dilemmas such as replay versus privacy, transfer versus interference, and task-aware versus task-agnostic learning reflect practical tensions that researchers must address when designing CL systems. Understanding these opposing forces is critical for developing models that can learn continuously over time without succumbing to either forgetting or intransigence.

\begin{figure*}[t]
  \centering
  \includegraphics[trim={0.5cm 0.5cm 0.5cm 0.5cm},clip,width=\textwidth]{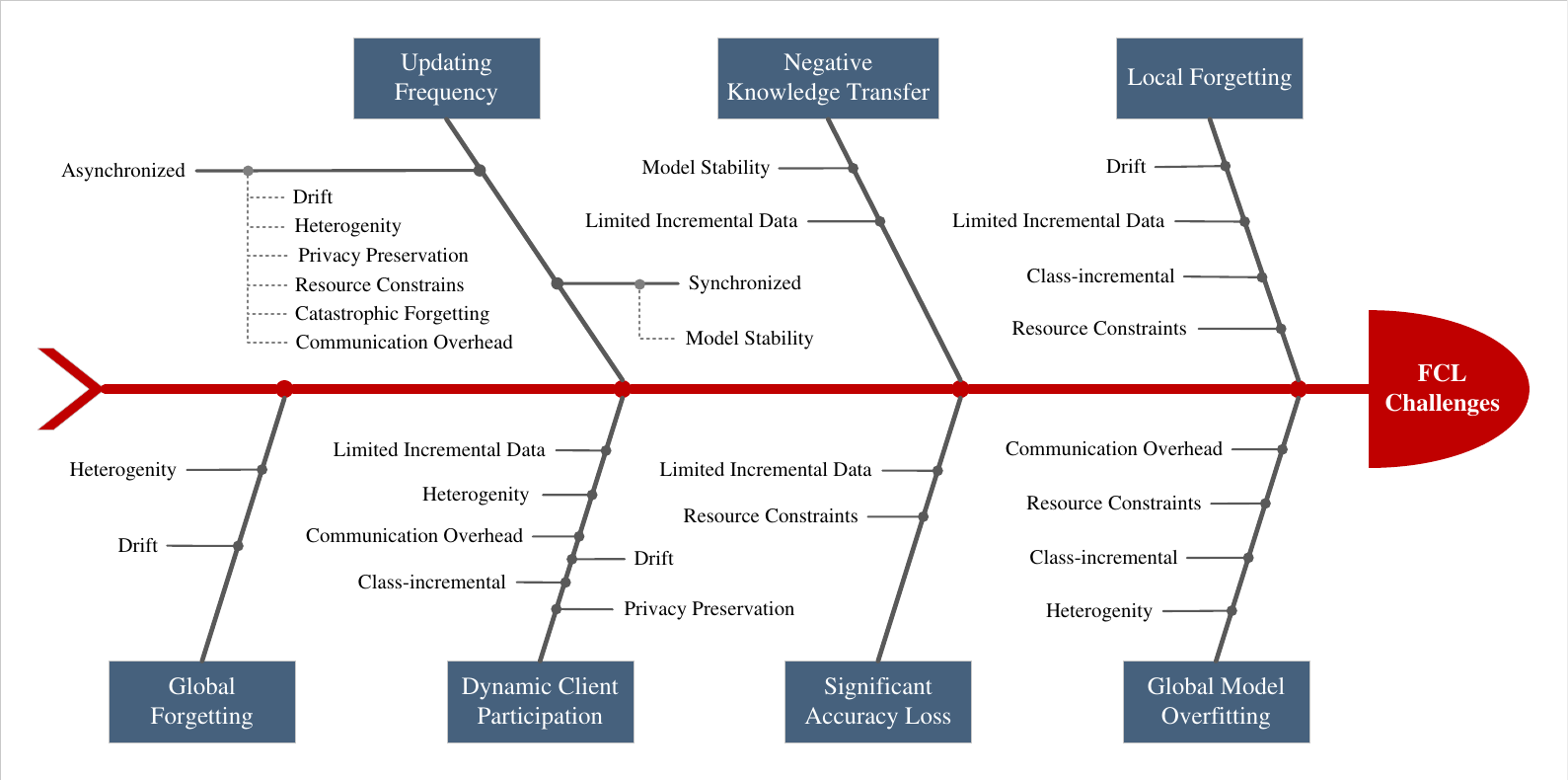}
  \caption{Fishbone diagram showing the taxonomy of challenges in FCL. Main branches show the main categories of these challenges, and sub-branches indicate specific reasons for these challenges within the specified domain.}
  \label{fig:fcl_challenges}
\end{figure*} 

\section{Challenges of Federated Continual Learning}
\label{sec:FCL}

In an FCL environment, numerous devices send their updates to a global model, necessitating continuous updates due to incoming data. Devices incrementally send the learned information to the global model, introducing specific challenges unique to the FCL environment. These challenges occur locally, globally, and during the transfer of knowledge between the global and local models. Managing incoming data locally, while keeping the global model updated presents significant difficulties, underscoring the importance of addressing these global model challenges.

\begin{table*}[t!]
\centering 
\footnotesize
\setlength{\tabcolsep}{3pt}
\caption{Main FCL approaches. Base models: Convolutional Neural Network (CNN), Recurrent Neural Network (RNN), Multi-Layer Perceptron (MLP), Reduced MobileNet-V2 (RMNv2) \cite{9031131}, Graph Convolutional Network (GCN), Deep Graph Library (DGL), Deep Deterministic Policy Gradient (DDPG), Fully Connected Layers (FCLs), Model-Agnostic Meta-Learning (MAML), Attention-Based Model (Att), and Vision Transformer (ViT). Aggregators: Elastic Averaging SGD (EASGD) \cite{NIPS2015_d18f655c}, Prototypes Aggregator (PA), Weighted Ensemble-Based Aggregation (WEA), projection in turn (PIT), gradient information-based aggregation (GIA), Digital Twin Network (DTN), Gradient Integrator (GI), Batch-oriented Aggregation (BoA), Coordinate-based Model Aggregation (CooMA), and Adaptive Aggregation (AdapA). Learning strategies: supervised (S), unsupervised (U), and semi-supervised (SS). Updating: synchronous (Sync) and asynchronous (Async). Client participation: static (Stc) and dynamic (Dyn). FL architecture: centralized (Cent) and decentralized (Dcent). Streaming strategies: sliding window (SW), time window (TW), convolution kernel (CK), coreset selection(CS), adaptive windowing (ADWIN), reservoir sampling (RS), and data buffer (DB). FL Challenges: communication overhead (CO), model stability (MS), heterogeneity (H), resource constraints (RC), and privacy preservation (PP). CL challenges: concept drift (CD), class incremental learning (CI), limited incremental data (LD), and catastrophic forgetting (CF). FCL challenges: global forgetting (GF), global model overfitting (GMO), negative knowledge transfer (NT), significant accuracy loss (AL), and local forgetting (LF). * and \ding{55} denote not applicable and not mentioned.}

    \resizebox{\textwidth}{!}{
    \begin{NiceTabular}{lccccccccccc} 
    \toprule
    &\multicolumn{8}{c}{Topology/Techniques}&\multicolumn{3}{c}{Challenges}\\
    \cmidrule(l){2-9}
    \cmidrule(l){10-12}
    
    Approaches & 
    \makecell{Model\\Homogeneity} &\makecell{Base\\Models} & \makecell{Client\\Participation}&
    
    \makecell{FL\\Architecture} & Updating & Aggregator  & \makecell{Learning\\Strategy}  & \makecell{Stream\\Handling}& 
    
    FL&CL&FCL\\ 
    
    \midrule
    
    Fed-ReMECS \cite{NANDI2022340}&
    \checkmark&FFNN&Stc&Cent&Sync&Fedavg&S&SW&PP&\textasteriskcentered&\textasteriskcentered\\
    
    ISCN \cite{10294732}
    &\checkmark&\textasteriskcentered&Stc&
    Cent&\textasteriskcentered&ISCN&
    S&SW&
    PP&LD&\textasteriskcentered\\
    
    \cite{app121910025}
    &\checkmark&SE+ATT&Stc&
    Cent&\textasteriskcentered&FedAvg&
    S&CK&
    H&\textasteriskcentered&\textasteriskcentered\\

    FedSpace \cite{10208460}
    &\checkmark&ResNet&Stc&
    Cent&Async&FedAvg&
    S&\textasteriskcentered
    &MS&CL&GF\\

   FedConD \cite{9671924}
   &\textasteriskcentered&\textasteriskcentered&Dyn
   &Cent&Async&FedConD&
   S&\textasteriskcentered&
   CO/MS&CD&\textasteriskcentered\\

   ASO-Fed \cite{9378161}&
   \textasteriskcentered&\textasteriskcentered&Dyn&Cent&
   Async&ASO-Fed&S&
   \textasteriskcentered&H/CO&
   CD&\textasteriskcentered\\

   \cite{9562751}&
   \ding{55}& BGD/SGD/mini-batch SGD&
   Dyn&Cent&Async&FedAvg&
   S&strSAGA\cite{NEURIPS2018_cebd648f}&
   MS/RC&\textasteriskcentered&\textasteriskcentered\\

   CHFL \cite{9892815}&
   \textasteriskcentered&\textasteriskcentered&Stc&
   Cent&Sync&FedAvg/VFL
   &S&\textasteriskcentered
   &MS/CO&CD&LF\\

   Cross-FCL \cite{9960821}&
   \checkmark&\textasteriskcentered&Dyn/Stc&
   Cent&Sync&FedAvg/FedAvg+EWC&
   S&Regularization Term&
   H/CO&CD&NT\\

   FedDrift-Eager/FedDrift \cite{pmlr-v206-jothimurugesan23a}&
   \checkmark&\textasteriskcentered&Stc&
   Dcent&Sync&FedAvg&
   S&SW&
   H&CD&\textasteriskcentered\\

   Master-FL \cite{10198306}&
   \textasteriskcentered&\textasteriskcentered&Stc&
   Cent&Sync&FedAvg/FedOMD\cite{9683589}&
   S&\textasteriskcentered&
   H&CD&\textasteriskcentered\\

   SOFed \cite{10128673}&
   \ding{55}&BYOL/SimCLR/MoCo&Dyn&
   Cent&Sync&FedAvg&U&
   CS&
   H&CF&LF\\

   FLARE \cite{10182870}&
   \checkmark&CNN&Stc&
   Cent&Sync&FedAvg&S&
   TW&
   CO&CD&\textasteriskcentered\\
   
   \cite{NEURIPS2023_d160ea01}&
   \checkmark&CNN&Stc
   &Cent&Sync&FedAvg&S&
   \textasteriskcentered&
   PP&CF&GF/LF\\

    FairFedDrift \cite{salazar2024unveilinggroupspecificdistributedconcept}&
    \checkmark&CNN&Stc
    &Cent&Sync&FedAvg&S&
    \textasteriskcentered&
    MS&CD&\textasteriskcentered\\

    \cite{hu2024energyefficientfederatededgelearning}&
    \checkmark&CNN&Dyn&
    Cent&Sync&FedAvg&
    S&\textasteriskcentered&
    CO/RC& CD&\textasteriskcentered\\
    
    ADMM-FedMeta \cite{10.1145/3466772.3467038}&
    \checkmark&CNN&Stc&
    Cent&Async&Inexact-ADMM&
    S&\textasteriskcentered&
    MS/RC&LD/CF&LF\\
    
    FedINC \cite{10.1145/3625687.3625800}&
    \checkmark&CNN/ResNet&Dyn&
    Cent&Sync&\makecell{FedAvg+PA}&
    S&\textasteriskcentered&
    H&CD/LD/CF&GF/LF\\

    Adaptive-FedAVG \cite{9533710}&
    \checkmark&MLP/CNN&Stc&
    Cent&Sync&Adaptive-FedAVG&
    S&\textasteriskcentered&
    H&CD/CI&\textasteriskcentered\\

    AFAFed \cite{BACCARELLI2022376}&
    \checkmark&IoTNet \cite{s19245541}/RMNv2&Stc&
    Cent&Async&EASGD&
    S&\textasteriskcentered&
    CO/H/RC&\textasteriskcentered&\textasteriskcentered\\

    O-GFML/PSO-GFML \cite{9751160}&
    \ding{55}&\textasteriskcentered&Dyn&
    Dcent&Sync&FedAVG&
    S&\textasteriskcentered&CO/RC&CD&\textasteriskcentered\\

    GLFC \cite{Dong_2022_CVPR}&
    \checkmark&ResNet \cite{He_2016_CVPR}&Dyn&
    Cent&Sync&\textasteriskcentered&
    S&\textasteriskcentered&
    pp&CI/CF&GF/LF\\

    \cite{10193322}&
    \checkmark&\textasteriskcentered&Stc&
    Cent&Sync&FedAVG&
    S&TW&
    CO/RC&CD&\textasteriskcentered\\

    CoOptFL/DYNAMITE \cite{10330725}&
    \checkmark&SVM/CNN&Stc&
    Cent&Sync/Async&\textasteriskcentered&
    S&RS&
    H/RC&CD&\textasteriskcentered\\
    Fed-IWDS \cite{electronics11223668}&
    \checkmark&CNN&Dyn&
    Cent&Async&FedAvg&S&\textasteriskcentered&
    H/RC/PP&CD&\textasteriskcentered\\

    \cite{electronics12040894}&
    \checkmark&CNN&Stc&
    Cent&Sync&S&\textasteriskcentered&\textasteriskcentered&&&\\

    FedSKF \cite{electronics13091772}&
    \checkmark&ResNet/VGG&Stc&
    Cent&Sync&PIT&S&\textasteriskcentered&
    H/PP & CI/CF& GF/LF\\

    \cite{10470505}&
    \checkmark&ResNet/CNN&Stc&
    Cent&Async&FedAvg \cite{pmlr-v54-mcmahan17a}&S&\textasteriskcentered&
    RC&CD&\textasteriskcentered\\

    \cite{10333463}&
    \checkmark&CNN&Stc&
    Cent&Sync&FedProx \cite{MLSYS2020_1f5fe839}&S&\textasteriskcentered&
    PP&\textasteriskcentered&\textasteriskcentered\\

    FCIL-MSN \cite{10540639}&
    \checkmark&ResNet&Dyn&
    Cent&Sync&GIA&S&\textasteriskcentered&
    CO/MS/H&CD/CF&GF/LF\\

    FedADC \cite{9517850}&
    \checkmark&CNN&\textasteriskcentered&
    Cent&Sync&SLOWMO \cite{DBLP:journals/corr/abs-1910-00643}&S&\textasteriskcentered&
    MS/H&CD&\textasteriskcentered\\

    FCL4SR \cite{10309661}&
    \checkmark& GCN+DGL&Dyn&
    Dcent&Sync&Elastic Transfer&
    S&\textasteriskcentered&
    PP&CD/CF&GF/NT/LF\\

    \cite{9821057}
    &\checkmark&CNN \cite{9439129}&Stc&
    Cent&Sync&FedAvg&S&\textasteriskcentered&
    RC&CF&LF\\

   FCL-SBLS \cite{10143925}&
   \checkmark&DDPG-based algorithm&Dyn&
   Cent&Sync&DTN&S&\textasteriskcentered&
   PP&CF&GF/LF\\

   FedStream \cite{10175385}&
   \textasteriskcentered&\textasteriskcentered&Stc&
   Cent&Sync/Async&FedAvg&S&DB&
   CO/H/RC&CD&\textasteriskcentered\\

   FedStream \cite{10198520}&
   \checkmark&K-Means+KNN \cite{ROSEBERRY202110}&Stc&
   Cent&Sync&FedAvg&S&\textasteriskcentered&
   CO/PP&CD/CI&\textasteriskcentered\\

   \cite{10406164}&
   \checkmark&CNN+FCLs&Stc&
   Cent&Sync&FedAvg&S&\textasteriskcentered&
   PP&CI/CF&GF/LF\\

   FedProK \cite{Gao_2024_CVPR}&
   \checkmark&ResNet&Stc&
   Cent&Sync/Async&FedAvg&S&PKF/FT&
   H&CI/CF&GF\\

   FedKNOW \cite{10184531}&
   \checkmark&ResNet/CNN&Stc&
   Cent&Sync&FedAvg&S&\textasteriskcentered&
   CO&CF&NT/LF\\

   FedRCIL \cite{Psaltis_2023_ICCV}&
   \checkmark&ResNet&Stc&
   Cent&Sync&FedAvg&S&DB&
   H/CO	&CF	&GF\\
   
   LCFIL \cite{9973580}	&
   \checkmark&MLP/LeNet/VGG/AlexNet&Dyn&Cent&Sync&S&\textasteriskcentered&
   MS&&&\\

   Re-Fed \cite{Li_2024_CVPR}&
   \checkmark&ResNet&Dyn&Cent&Sync&FedAvg&S&\textasteriskcentered&
   H	&CF&GF/LF\\

   ICMFed \cite{math11081867}&\checkmark&DenseNet/EfficientNet&Dyn&Cent&Sync&ICMFed+FOMAML/ICMFed+Reptil&S&\textasteriskcentered&
   CO/H&CD/CF&LF\\

   LGA \cite{10323204}&\checkmark&ResNet&Dyn&Cent&Sync&
   FedAvg&S&DB&
   H&CI/CF&GF/LF\\

   \cite{10295979}&\textasteriskcentered&\textasteriskcentered&
   Stc&Cent&Sync&FedAvg&S&FIFO/SRSR/DRSR&MS/RC&CD&\textasteriskcentered\\

  Flash \cite{pmlr-v202-panchal23a}&
  \checkmark&FCN/LSTM/CNN/ResNet&Dyn&Cent&Sync&Flash&S&\textasteriskcentered&
  H&CD&\textasteriskcentered\\

  \cite{10097140}&
  \checkmark&BERT-base model \cite{devlin-etal-2019-bert}&Dyn&Cent&Sync&FedAvg&S&TW&
  RC	&CF&	LF\\

  FedPC \cite{Yuan_2023_CVPR}&\checkmark&ResNet&Stc&Dcent&\ding{55}&\ding{55}&S&\textasteriskcentered&CO/RC/PP&CD/CF&LF\\

  cTD-$\alpha$MAML \cite{10148063}&\checkmark&MAML&Stc&Cent&Sync&FedAvg&S&\textasteriskcentered&
  CO/RC/PP&CD/CF&LF\\

  SFLEDS \cite{MAWULI2023119235}&
  \checkmark&KNN&Stc&Cent&Sync&KNN+Selected Prototypes&SS&DB&
  PP&CD/LD&\textasteriskcentered\\
  
  FedNN \cite{KANG2024110230}&
  \checkmark&LeNet \cite{726791}/ResNet&Stc&Cent&Sync&FedDC+FedNN&S&\textasteriskcentered&CO/MS/H&CD&\textasteriskcentered\\

  Fed-SOINN \cite{ZHU2022168}&
  \checkmark&Att&Dyn&Cent&Sync&FedAvg&SS&\textasteriskcentered&CO/MS&\textasteriskcentered&\textasteriskcentered\\
  RRA-FFSCIL \cite{JIANG2024127956}&
  \checkmark&ResNet&Dyn&Cent&Sync&\textasteriskcentered&S&\textasteriskcentered&H&LD/CF&GF/LF\\

  \cite{9950044}&
  \checkmark&VPT360 \cite{9733647}&Stc&Cent&Sync&FedAvg&S&\textasteriskcentered&
  H/PP&CF&LF
  \\
  \cite{YANG202416}&\checkmark&Multi-head AlexNet&Stc&Cent&Sync&CooMA&S&\textasteriskcentered&
  H/RC & CF& GF/LF\\

  \cite{LI2024111491}&
  \checkmark&LSTM+FC&Stc&Cent&Sync&AdapA&S&SW&
  H&CD/CF&GF/LF\\

  FL-IIDS \cite{JIN202457}&
  \checkmark &CNN-GRU&Dyn&Cent&Sync&FedAvg&S&\textasteriskcentered&MS/RC&CD/CF&
  GF/LF\\

 FedViT \cite{ZUO20241}&\checkmark&ViT&Stc&Cent&Sync&FedAvg/FedKNOW&S&\textasteriskcentered&CO/MS&CD/CF&GF/NT/AL\\

 CFeD \cite{ijcai2022p303}&\checkmark&TextCNN \cite{kim-2014-convolutional}/Resnet+FC&Dyn&Cent&Sync&CFeD&S&\textasteriskcentered&\textasteriskcentered&CF&
 GF/LF\\

 FedET \cite{liu2023fedetcommunicationefficientfederatedclassincremental}&\checkmark&ResNet/ViT-Base& Dyn&Cent&Sync&\textasteriskcentered&S&\textasteriskcentered&CO&CI/CF&GF/LF\\

 FedWeIT \cite{pmlr-v139-yoon21b}&
 \checkmark&LeNet/ResNet&Stc&Cent&Sync& EWC method&S&\textasteriskcentered&
 CO&CF&NT/LF\\
 
 \cite{good2023coordinatedreplaysampleselection}&\checkmark&TinyBERT \cite{jiao-etal-2020-tinybert}&Dyn&Cent&Sync&FedAvg&S&\textasteriskcentered&RC&CF&LF\\
 
 \cite{hendryx2021federatedreconnaissanceefficientdistributed}&\checkmark&4-Conv/Resnet&Stc&Cent&Sync&FedAvg&S&\textasteriskcentered&CO/RC	&CF&	LF\\

\cite{10543076}&\checkmark&		Conv+FC+LSTM&Stc&Cent&Sync&FedAvg&S&\textasteriskcentered&PP&\textasteriskcentered&\textasteriskcentered\\

SLMFed \cite{10399971}&\checkmark&CNN&Dyn&Cent&Sync&SLMFed&S&\textasteriskcentered&H&CF&LF\\		

TARGET \cite{10376970}&\checkmark&ResNet&Dyn&Cent&Sync&Target&S&\textasteriskcentered&PP&LD/CF&GF/LF\\

HFIN \cite{10546981}&\checkmark&1D-CNN&Stc&Dcent&\ding{55}&\ding{55}&S&\textasteriskcentered&\textasteriskcentered&CF&LF\\

 \cite{CHAVES2024101036}&\textasteriskcentered&Varied&Stc&Cent&Async&Varied&S&\textasteriskcentered&H&LD&\textasteriskcentered\\

\cite{qi2023bettergenerativereplaycontinual}&\checkmark&CNN&Stc&Cent&Sync&FedCIL&S&\textasteriskcentered&H/MS&CI/CF&GF/LF\\

 FedET\cite{liu2023fedetcommunicationefficientfederatedclassincremental}&\checkmark&ResNet/ViT/Bert-Base-Uncased\cite{wolf-etal-2020-transformers}&Dyn&Cent&Sync& Weighted Averaging(distillation)&S&\textasteriskcentered&H&CF&GF/LF\\

FBL\cite{Dong_2023_CVPR}&\checkmark&Deeplab-v3\cite{7913730} with ResNet&Dyn&Cent&Sync&\textasteriskcentered&S&\textasteriskcentered&H&CI/CF&GF/LF\\

PP-FCIL\cite{10363211}&\checkmark&ResNet&Dyn&Cent&Sync&FedAvg&S&\textasteriskcentered&PP&CI/CF&LF\\

\cite{wisdom}&\textasteriskcentered&\textasteriskcentered&Dyn&Cent&Sync&FedAvg&SS&\textasteriskcentered&MS/H/PP&LD/CF&GF/GMO/AL\\

FedMAC \cite{fusion}&\ding{55}&\makecell{MobileNetV2\cite{howard2017mobilenetsefficientconvolutionalneural}\\/MobileBERT\cite{sun2020mobilebertcompacttaskagnosticbert}}&Dyn&Cent&Sync&\makecell{FedAvg/FedProx\\/FedRS\cite{10.1145/3447548.3467254}/FedOpt\cite{app10082864}}&S&\textasteriskcentered&H&\textasteriskcentered&\textasteriskcentered\\

EvoFedIDS\cite{Industrial}&\checkmark&CNN&Dyn&Cent&Sync&FedAvg&S&\textasteriskcentered&H/RC&CF&GF/LF\\

\cite{Diffusion}&\checkmark&ResNet&Stc&Cent&Sync&FedAvg&S&\textasteriskcentered&H/PP&CF&LF\\

FedMGP\cite{Personalized}&\checkmark&ResNet&\textasteriskcentered&Cent&Sync/Async&FedMGP&S&\textasteriskcentered&H/PP&CF&GF/LF\\

FCLLM-DT\cite{FCLLM}&\checkmark&CNN&Stc&Cent&Sync&FedAvg&S&SW&H/PP&LD&\textasteriskcentered\\

FCL4DD\cite{Weakly}&\checkmark&ResNet&Stc&Cent&Sync&FedAvg&S&\textasteriskcentered&H&CF&GF/LF\\

\cite{10945356}&\checkmark&ResNet&Stc&Dcent&\ding{55}&\ding{55}&S&\textasteriskcentered&CO/PP&CI/CF&AL/LF\\

MeCo\cite{10899876}&\checkmark&ResNet&\textasteriskcentered&Cent&Sync&FedAvg&S&\textasteriskcentered&CO/MS/H/RC&CF&GF/LF\\

Loci\cite{10857343}&\ding{55}&Varied&Dyn&Cent/Dcent&Sync&FedAvg&S&\textasteriskcentered&CO/H&\textasteriskcentered&NT\\

SacFL\cite{11006111}&\checkmark&LeNet/ResNet/TextCNN&Dyn&Cent&Sync&FedAvg&S&\textasteriskcentered&H/RC&CD&\textasteriskcentered\\

     \bottomrule
    \end{NiceTabular}
}

\label{tab:approaches}
\end{table*}

As illustrated in Figure \ref{fig:fcl_challenges}, FCL challenges are categorized into three main areas: local challenges, global challenges, and challenges related to knowledge transfer between global and local models. Each of these categories is further divided into sub-challenges, which are discussed in the following sections. Their origins (root causes) and the proposed solutions are examined in detail, emphasizing the complexity and importance of overcoming these obstacles to achieve effective FCL.

\subsection{Global Model Challenges}

One of the main concerns in FCL  is how clients can handle continual data, while simultaneously learning in a distributed manner. It is critical to understand how the challenges faced by clients affect global training and how these challenges can be tackled. In this section, four main challenges are discussed: significant accuracy loss, catastrophic forgetting, global model overfitting, and negative knowledge transfer. The solutions to these challenges are also explained. Table \ref{tab:approaches} summarizes approaches to address global model overfitting and catastrophic forgetting in FCL.

\subsubsection{Catastrophic Forgetting}

Catastrophic forgetting, also known as global forgetting, is a significant challenge in FCL, caused by class imbalance and data heterogeneity among clients in the FL  environment, and class-incremental and limited incremental data issue in the CL  environment. This leads the global model to adapt to dominant classes or new data distributions, often forgetting earlier tasks or less frequent classes. Also, training with a small amount of labeled data not only can deteriorate the model performance on new tasks, but also exacerbate catastrophic forgetting of previous knowledge. To address this, strategies are needed to preserve and transfer new client information despite limited resources and communication bandwidth in FL environments. Tables \ref{tab:evaluation} and \ref{tab:dataset} list datasets supporting class-incremental and instance-incremental tasks relevant to these strategies. As depicted in Figure \ref{fig:forgetting}, three primary methods to mitigate global forgetting are preserving the global model, learning from exemplars, and a combination of these two.

To preserve the global model in FCL, knowledge fusion is used to leverage the previously trained global model, integrating new and old data effectively. Five primary methods are employed for this purpose: proxy server methods, prototypical learning, optimization and Loss function modification, knowledge distillation, and gradient-based methods.

Proxy server methods utilize intermediary servers to store and manage previous global model data. For instance, \cite{10323204} selects the best old model for global anti-forgetting and augments new prototype images via self-supervised prototype augmentation. Similarly, \cite{Dong_2022_CVPR} proposes a proxy server that selects the best old global model to assist local relation distillation.

Prototypical Learning focuses on representative examples to maintain class prototypes. For example, \cite{Gao_2024_CVPR} uses prototype vectors to average client model representations, preserving knowledge across tasks.

Adjusting optimization techniques and loss functions balance new and old knowledge. This includes loss function modifications, learning rate adjustments, and proximal terms. \cite{10208460} introduces a specific loss function to distill knowledge from the old global model, employing augmented old global prototypes and a contrastive representation loss to properly allocate new global class prototypes and align new local classes with their previous positions. Yuan et al. \cite{Yuan_2023_CVPR} mitigate catastrophic forgetting by adjusting the learning rate over time.

Knowledge distillation is a widely used technique for fusing knowledge in FCL \cite{usmanova2021distillationbasedapproachintegratingcontinual, liu2023fedetcommunicationefficientfederatedclassincremental}. The core idea involves transferring the knowledge from an already trained teacher model to a student model. Essentially, the teacher model guides the student to replicate the same mapping relationship by providing the label $Y$ for the sample $X$, similar to the way humans teach. The continual distillation mechanism uses distillation techniques at multiple scales and employs contrastive learning to transfer knowledge from old tasks to new tasks while preserving the learned knowledge \cite{Psaltis_2023_ICCV}. Federated Not-True Distillation (FedNTD) \cite{NEURIPS2022_fadec8f2} focuses on preserving the global model's knowledge by only considering not-true classes during distillation, thus maintaining the global perspective without directly accessing local data. The Enhancer distillation method \cite{liu2023fedetcommunicationefficientfederatedclassincremental} incorporates a small module, the enhancer, to absorb and communicate new knowledge, thereby mitigating the imbalance between old and new knowledge. Lastly, exemplar-free distillation \cite{JIN202457} avoids the use of a rehearsal buffer by employing distillation methods that do not require storing past data samples. In the context of a decentralized federated learning framework for joint class and domain continual learning, \cite{10945356} employs response-based knowledge distillation by aligning the logit vectors of local models with those of an incrementally updated global model. This approach mitigates catastrophic forgetting without requiring access to data from previous nodes.

Moreover, some papers prevent the forgetting of previously learned tasks by modifying gradients. Accordingly, \cite{10184531} integrates gradients from the current task with gradients from the most dissimilar previous tasks, as similar as \cite{10323204}, which adjusts gradient compensation dynamically to balance the forgetting rates of different categories, ensuring that no single category is disproportionately forgotten.

\paragraph{Learning From Exemplars}
In contrast to the first approach, in which the previously learned information is preserved by fusing the previous global model and using only the data from the current task, an exemplar refers to a sample from a previous task that is stored in a memory buffer for future training that is another approach for preserving previous information. There are different methods for selecting a small portion of previous data as an exemplar, which fall into two main sections, global exemplar and local exemplar. The global exemplar approach involves the server aggregating a subset of data from clients and redistributing it to clients during training. However, this method does not adhere to the principles of FL. On the other hand, local exemplar refers to each client retaining a subset of data from previous tasks for use in future training. On the other hand, local exemplar refers to each client retaining a subset of data from previous tasks for use in future training.

Replay-based methods refer to approximating and recovering old data distributions, which is the principle way for learning from local exemplars in FCL and are divided into two sub-directions depending on the content of replay, direct and generative replay. Direct replay, known as experience replay, involves storing and replaying actual data samples from previous tasks during the training of new tasks. Replay-based methods can be optimized using several strategies such as importance-based selection, coordinated replay, split-based, and retainer mechanisms. For instance, \cite{Li_2024_CVPR} selectively stores the most critical data samples to ensure that essential knowledge is retained. This work also considers coordination among clients for better sample selection, ensuring that important samples are replayed. Also, \cite{YANG202416} utilizes a retainer mechanism to ensure that valuable information from previous tasks is preserved. On the other hand, generative replay (pseudo-rehearsal) uses generative models to synthesize data samples that approximate the distribution of previous tasks. In this context, \cite{NEURIPS2023_d160ea01} proposes server-side generative model training, in which the server trains a generative model in a data-free manner at the end of each task. This model synthesizes images that mimic the distribution of old tasks, which can then be used to update the global model. This helps the global model retain knowledge from all previous tasks, thus mitigating global forgetting. 

These approaches help in maintaining a balance between acquiring new information and retaining previously learned knowledge, and, thus, mitigating global forgetting in FCL.

\subsubsection{Global Model Overfitting}

Global model overfitting in FCL  is a significant challenge that occurs when the global model, which is aggregated from the local models of different clients, performs exceptionally well on the training data (often non-IID and limited to specific clients) but fails to generalize to unseen or new data. Several factors contribute to this problem: resource constraints, communication overhead, heterogeneity, which are rooted in FL, class-incremental, and limited incremental data that arise due to the unique characteristics of CL. Resource constraints, such as limited computational power and memory on client devices, force the use of simplified models or reduced training iterations, which can lead to underfitting at the local level but overfitting at the global level, when these simplified models are aggregated. Communication overhead limits the frequency of model updates and necessitates the use of compression techniques that can degrade model quality, leading to overfitting as the global model may overly adapt to the noisier, and compressed updates from clients. Heterogeneity among clients, with varying data distributions and device capabilities, causes the global model to struggle with generalization, as it may overfit to the dominant clients' data distributions, neglecting the broader population. In CL, class-incremental tasks, where new classes are introduced over time, further complicate the global model's ability to generalize, as it may overfit to recent classes at the expense of forgetting previous ones. Finally, limited incremental data due to the sequential nature of CL forces the model to deal with sparse data, increasing the likelihood of overfitting to these small and biased datasets rather than learning a robust and generalized representation. These interconnected challenges highlight the complexity of preventing overfitting in FCL, where solutions must address the interplay between FL's distributed nature and CL's dynamic task environment. 
Overfitting may accur in two cases, when the global model focuses on the past data and overfited on them, which causes low performance and not learning the new data. On the other hand it can be that the global model is overfitted on new data, which causes catastrophic forgetting. So, the core idea of solutions is to make a balance between learning new data and also preserving previously learnt information. These solutions are categorized into five categories, traditional methods, optimization, balancing learning, generalization, and few-shot learning \cite{JIANG2024127956, Wang2020}. To enhance the robustness of a machine learning model against overfitting, several key factors should be considered. These principles are equally applicable when designing the global model in FCL  systems. By carefully addressing these factors, the model's ability to generalize to new data can be significantly improved, ensuring better performance across diverse environments and reducing the risk of overfitting.

\begin{figure}[t]
  \centering
  \includegraphics[trim={0.6cm 0.6cm 0.6cm 0.6cm},clip,width=\columnwidth,page=2]{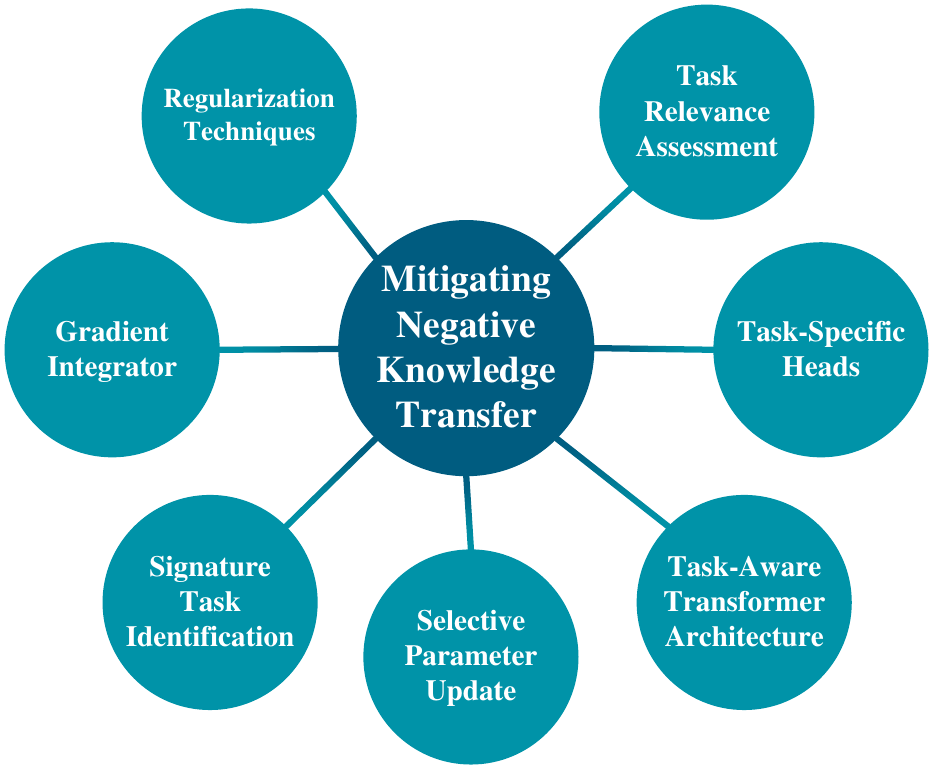}
  \caption{Categorization of techniques used to mitigate negative knowledge transfer}
  \label{fig:transfer}
\end{figure}

Solutions to FCL typically require an elaborate balance of learning strategies, updating mechanisms, and aggregation techniques aimed at dealing with the unique challenges of distributed and dynamic environments. These approaches often fall under supervised, unsupervised, and semi-supervised learning, with model variants ranging from CNNs and RNNs to state-of-the-art transformers such as Vision Transformers (ViTs). Aggregators such as Elastic Averaging SGD and weighted ensemble-based methods provide additional model stability and adaptability. These are all summarized in Table \ref{tab:approaches}, which explains the relationships among base models, aggregation strategies, and specific challenges faced by FCL, such as catastrophic forgetting, global model overfitting, and communication overhead. This table provides a detailed reference that facilitates indicating how various techniques meet different requirements of FCL systems.

\subsubsection{Negative Knowledge Transfer}

Negative knowledge transfer in FCL  refers to the phenomenon, where knowledge acquired from one task or dataset adversely affects the model's performance on subsequent tasks or datasets. Table \ref{tab:approaches} reports models and aggregation techniques that mitigate this issue. This issue often arises when the knowledge gained from previous tasks is either irrelevant or harmful to new tasks, leading to situations where averaging user-specific models actually degrades performance for certain subjects. In the context of FCL, where multiple clients (or edge devices) continuously learn from their local data and contribute to a global model, negative knowledge transfer can occur due to several factors, including data heterogeneity, model stability, limited incremental data, and catastrophic forgetting.

One of the primary causes of negative knowledge transfer in FCL is the heterogeneity of data across clients, a common challenge in FL. When clients have data that is not identically and independently distributed (non-IID), the resulting data distribution across clients can vary significantly. This variability can cause the global model to be influenced by irrelevant or conflicting knowledge from different clients, thereby diminishing its overall performance. Additionally, limited incremental data and issues with model stability can exacerbate negative knowledge transfer. Clients typically train their models on local data, which may be biased or not representative of the global data distribution. As a result, when the global model aggregates updates from all clients, it may overfit to these local idiosyncrasies, leading to poor generalization to new tasks or data. Furthermore, as the model learns new tasks, it may experience catastrophic forgetting, where previously learned tasks are forgotten, especially if the new tasks involve learning patterns that conflict with older ones. This form of negative transfer ultimately deteriorates the model’s performance on earlier tasks as it adapts to new ones. 

\begin{table*}[t]
    \footnotesize
    \centering
    \setlength{\tabcolsep}{3pt}
    \caption{Performance Comparison of FedKNOW and FedViT: Accuracy Improvement and Forgetting Rate Across Tasks and Clients on CIFAR-100}
    \label{tab:sync}
    \begin{tabularx}{\textwidth}{
        p{25mm}                               
        >{\centering\arraybackslash}X         
        >{\centering\arraybackslash}X         
        >{\centering\arraybackslash}X         
        >{\centering\arraybackslash}X         
        *{10}{>{\centering\arraybackslash}X}  
    }
        \toprule
        \multirow{4}{*}{Methods} 
        & \multicolumn{4}{c}{Number of Clients} 
        & \multicolumn{10}{c}{Number of Tasks} \\
        
        \cmidrule(lr){2-5}
        \cmidrule(lr){6-15}

        & \multicolumn{2}{c}{50 Clients} 
        & \multicolumn{2}{c}{100 Clients}
        & 1 & 2 & 3 & 4 & 5
        & 6 & 7 & 8 & 9 & 10 \\

        \cmidrule(lr){2-3}
        \cmidrule(lr){4-5}
        \cmidrule(lr){6-15}

        & Accuracy & FR 
        & Accuracy & FR
        & \multicolumn{10}{c}{Improvement (\%)}  \\

        \midrule
   
        FedKNOW \cite{10184531}  
        & 59.26 & 11.30 & 53.18 & 7.30 
        & 36.52 & 74.74 & 82.58 & 84.84& 88.69 & 94.87 & 92.40 & 95.52 & 98.72 & 97.75 \\

        FedViT \cite{ZUO20241}  
        & 62.08& 1.70& 54.25& 4.23 
        & 64.6 & 74.22 & 88.3 & 87.5 & 95.0 & 103.03 & 107.69 & 106.81 & 113.74& 110.66 \\

        \bottomrule
    \end{tabularx}
    \label{tabel:ScalabilityFedKnowFedViT}
\end{table*}

As shown in Figure \ref{fig:transfer}, addressing negative knowledge transfer in FCL involves strategies such as careful design of aggregation methods, regularization techniques to preserve important knowledge, and ensuring that the learning process remains robust to the differences in data distributions across clients. These strategies are illustrated in Figure \ref{fig:transfer}. To address this challenge, \cite{10184531, ZUO20241, 10857343} proposes FedKNOW, FedViT, and Loci respectively, which are client-side solutions for mitigating negative knowledge transfer and a novel concept of signature task knowledge to extract and integrate the knowledge of signature tasks that are highly influenced by the current task. FedKNOW is designed to continuously extract and integrate knowledge from signature tasks that significantly influence current tasks by assessing the relevance of tasks, when integrating knowledge. This assessment helps in filtering out knowledge from tasks that may not align well with the current task's objectives. Each client in FedKNOW consists of a knowledge extractor, a gradient restorer, and, crucially, a gradient integrator. During the training of a new task, the gradient integrator plays a key role in preventing catastrophic forgetting and mitigating negative knowledge transfer. It achieves this by effectively combining the signature tasks identified from past local tasks with relevant tasks from other clients, facilitated through the global model. On the other hand, FedViT introduces a task-aware Vision Transformer architecture that is capable of distinguishing between different tasks and adapting its learning process accordingly, which reduces the risk of negative knowledge transfer by ensuring that the model's focus remains on relevant features and patterns specific to each task, rather than indiscriminately applying knowledge from previous tasks that may not be applicable. Also, this paper utilizes the selective parameter update mechanism, which carefully chooses which parts of the model should be updated based on the new task. Moreover, The model in FedViT is designed with task-specific heads, which means that different tasks can have their own dedicated output layers. This approach reduces the likelihood of negative knowledge transfer by isolating the outputs for different tasks, ensuring that the learning from one task does not adversely affect the performance of another task. \cite{9960821} addresses negative knowledge transfer by enhancing collaboration among edge devices to share relevant knowledge, employing selective knowledge sharing to filter out harmful information, and maintaining task awareness to prioritize valuable knowledge. Additionally, it uses gradient adjustment mechanisms to ensure new knowledge integrates without conflicting with prior learning, thereby enabling more robust and accurate learning across distributed edge environments.

\begin{figure}[t]
  \centering
  \includegraphics[width=\columnwidth ,trim={0cm 0cm 0.6cm 0cm},clip]{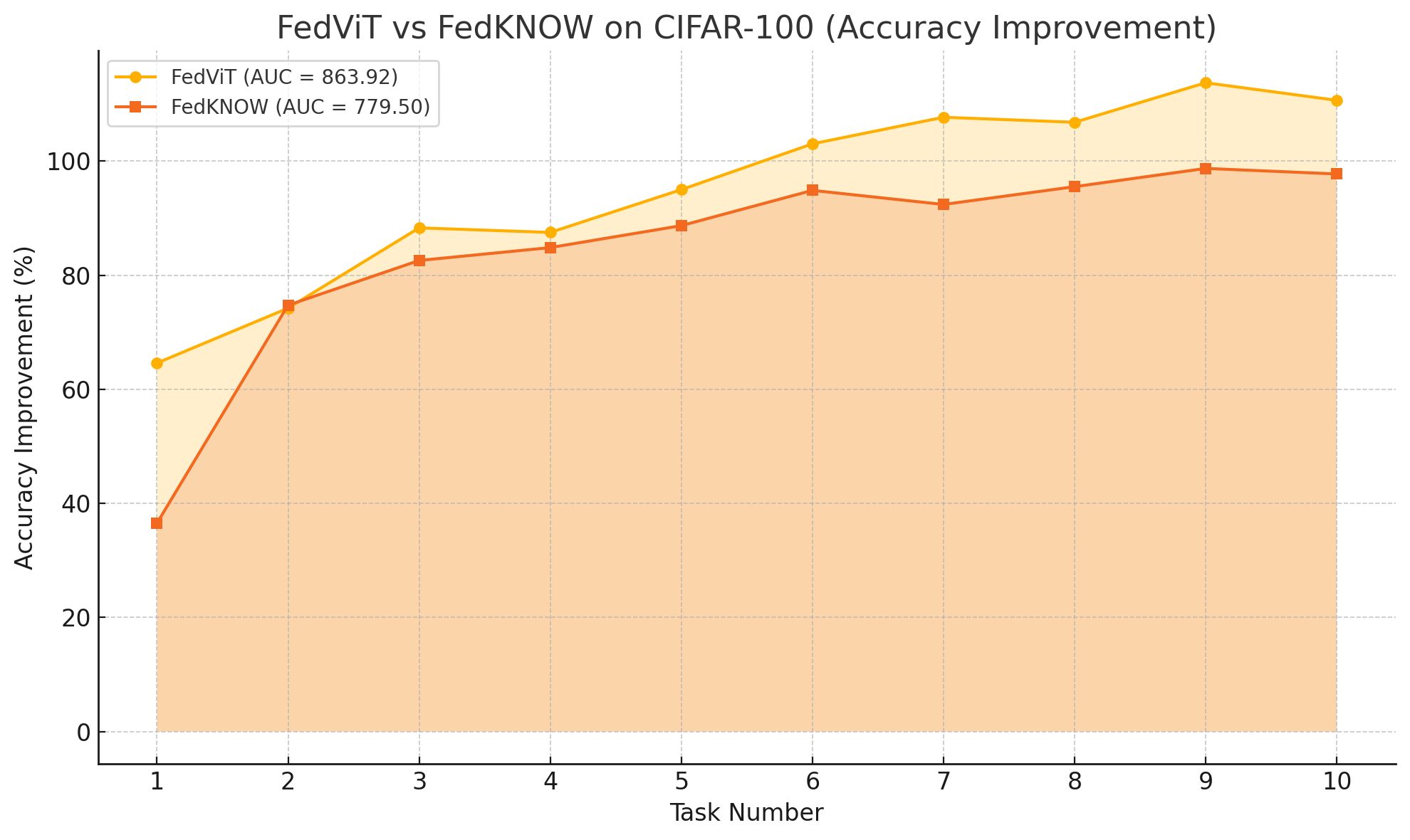}
  \caption{Cumulative Accuracy Improvement Comparison Between FedViT and FedKNOW on CIFAR-100: Area Under the Curve (AUC) comparison of average percentage accuracy improvement for FedViT and FedKNOW on the CIFAR-100 dataset across 10 incremental tasks. The AUC metric reflects the overall performance trend, with FedViT consistently achieving higher cumulative accuracy improvements, indicating stronger generalization and retention capabilities across tasks.}
  \label{fig:FedKnowFedViTAccuracy}
\end{figure}

Table \ref{tabel:ScalabilityFedKnowFedViT} and Figure \ref{fig:FedKnowFedViTAccuracy} compare the performance of FedKNOW and FedViT on the CIFAR-100 dataset across 10 tasks. The results show that FedViT consistently outperforms FedKNOW across all tasks, with notable accuracy improvements at each stage. On average, FedViT achieves a higher percentage of accuracy per task, indicating stronger generalization and better scalability to the number of tasks.
Furthermore, when evaluating scalability with respect to the number of clients (50 vs. 100), both methods experience a drop in accuracy. However, FedViT maintains higher accuracy overall, while the forgetting rate (FR) decreases in FedKNOW but increases slightly in FedViT. Despite this, the overall performance of FedViT remains superior in terms of both task and client scalability.

To compare computational overhead associated with FedViT and FedKNOW across the primary phases of knowledge extraction, storage, and gradient restoration. During knowledge extraction, FedViT introduces higher computational demands by evaluating loss values across the training dataset to identify high-quality samples, whereas FedKNOW employs a more efficient weight-based pruning strategy, retaining only a small fraction (10\%) of the most significant model weights. In terms of knowledge storage, FedViT maintains selected data samples, resulting in higher memory consumption, while FedKNOW achieves greater storage efficiency by preserving only the pruned model weights. Gradient restoration in FedViT requires computation over the retained sample subset, whereas FedKNOW reconstructs gradients using its stored weights in conjunction with data from the current task, reducing reliance on historical data. Both approaches employ quadratic programming to integrate gradients, incurring comparable computational complexity with polynomial time guarantees. Notably, FedViT mitigates training costs by freezing a portion of model parameters during training, while FedKNOW, despite not incorporating parameter freezing, demonstrates improved scalability and stability in training time as the retained knowledge size increases, highlighting its computational efficiency.

\subsubsection{Significant Accuracy Loss}
Many factors lead to significant accuracy loss in the global model. For instance, previuos challenges in the global model in FCL resulted in poor performance. However, it may caused by other factors such as model stability, which is rooted in FL and is one of the most prominent reasons. In general, the training procesure plays an important role in the model's capability to learn the main information, converge, and maintain good accuracy. Model complexity, hyperparameter tuning, and aggregation methods should be considered. Also, resource constraints in FL are another issue, which doesnt let the learnt information in local models aggregated completely so it cause a loss in final accuracy. Moreover, limited incremental data is another problem rooted in CL, which leads to a loss in acuuracy.

Several factors contribute to significant accuracy loss in global models, particularly in FL and CL  contexts. One of the primary challenges in FL is model stability, which can lead to unacceptable accuracy in the global model. This instability often stems from the distributed nature of FL, where the training process plays a crucial role in the model's ability to learn essential information, achieve convergence, and maintain high accuracy.

Moreover, model complexity, hyperparameter tuning, and aggregation methods are critical factors that influence the model's performance. In FL, resource constraints can hinder the effective aggregation of learned information from local models, further contributing to accuracy loss. Additionally, in CL, the availability of limited incremental data poses a significant challenge, leading to reduced model accuracy over time.

In summary, to enhance the accuracy of global models, it is essential to address issues related to model's stability, optimize training procedures, and carefully consider model complexity, hyperparameter tuning, aggregation methods, and resource constraints.
\subsection{Knowledge Dissemination}

 For streaming data, asynchronous updating might be favored to allow rapid adaptation, while synchronized updating might be preferred in static environments to ensure consistency across clients. To better understand how different update strategies influence learning in FCL, this section examines \textit{updating frequency} as a key challenge.

\begin{table}[t]
	\footnotesize
	\centering
	\setlength{\tabcolsep}{5pt}
	\caption{Advantages and disadvantages of synchronous and asynchronous updating in FCL.}
	\label{tab:sync}
	\begin{tabularx}{\columnwidth}{p{16mm}*{2}{>{\RaggedRight\arraybackslash} X }}
		\toprule
		Updates & Advantages & Disadvantages \\
		\midrule
		\multirow{3}{*}{Synchronous}
		& Consistency & Communication strains\\
		&Simplified aggregation & Scalability issues\\
		&Lower risk of staleness & -- \\
		\midrule
		\multirow{3}{*}{Asynchronous} 
		& Flexibility & Staleness of updates \\
		& Scalability & Complex aggregation \\
		& Fault Tolerance & --\\
		\bottomrule
	\end{tabularx}
\end{table}

In FCL, updating frequency refers to how often updates are communicated between the clients (distributed devices or nodes) and the central server. This concept is crucial in determining the efficiency, effectiveness, and overall performance of the learning system. There are two primary approaches to updating: synchronized and asynchronous. The comparison between synchronized and asynchronized updating is crucial for understanding the trade-offs in communication efficiency, model convergence, and overall system performance. Table \ref{tab:sync} illustrates the advantages and disadvantages of each method. 

In synchronized updating, all clients (or devices) train their local models on their respective data, and then, transmit the updated parameters to a central server simultaneously after a predetermined number of training iterations or epochs. The server aggregates these updates—often through methods such as averaging—to produce a global model, which is subsequently distributed back to the clients for the next round of training.

One of the key advantages of this approach is the consistency in updates, as all clients contribute to the global model simultaneously. This leads to more stable convergence and simplifies the aggregation process since all updates are received concurrently. Additionally, by using all updates at once, there is a reduced risk of outdated model parameters negatively impacting the global model.

However, synchronized updating also has its drawbacks. It necessitates that all clients be ready at the same time, which can introduce delays, particularly if some clients have slower networks or processing capabilities—a common scenario in FCL  due to the dynamic nature of incoming data. Moreover, the inherent flexibility of FCL allows for dynamic client participation, enabling the inclusion of new clients and the exclusion of existing ones as training progresses. This adaptability is crucial in real-world applications, where devices may connect and disconnect intermittently, data availability may fluctuate, or new data sources may emerge. As a result, synchronized updating can become time-consuming, as the process may experience significant delays due to the need to wait for all clients to be ready simultaneously. 

In asynchronous updating, clients transmit their updates to the server independently, without coordinating with other clients. The server integrates these updates into the global model immediately upon receipt, without waiting for all clients to complete their local training.

This approach offers several advantages. Clients can send updates as soon as they are ready, reducing idle times and enhancing overall system efficiency. Additionally, asynchronous updating allows the system to scale more effectively, as it does not require synchronization across a large number of clients. The system also demonstrates greater resilience to client dropouts or delays, as it does not rely on all clients being synchronized. From a practical standpoint, \cite{10208460} introduce an asynchronous federated continual learning framework in which clients learn tasks in different orders and at varying times. Their proposed method, FedSpace, tackles key challenges such as catastrophic forgetting and task misalignment, showing improved performance in asynchronous learning environments.

However, this method presents certain challenges. Updates may be based on outdated versions of the global model, potentially introducing inconsistencies and slowing convergence. The server must also manage and integrate updates that arrive at different times, which increases complexity and may necessitate more sophisticated algorithms to ensure stability.

To compare these two approaches, some works highlight that while synchronized updating ensures consistent model updates, it can increase latency due to stricter communication requirements. This is while asynchronized updating offers greater flexibility and faster updates; however, it may lead to inconsistencies in the aggregated model. Each study compares these methods across different aspects such as accuracy, efficiency, robustness, scalability, and practicality in their respective contexts \cite{Gao_2024_CVPR, pmlr-v139-yoon21b, dai_addressing_2023, BACCARELLI2022376}.

Accordingly due to the dynamic environment of FCL, the choice between synchronized and asynchronous updating in FCL  depends on the specific use case, the nature of the data distribution, and the system's constraints. Synchronized updating may be preferable in environments where consistency and stable convergence are critical, especially with static datasets, as it ensures that all clients update the global model based on the same version, maintaining accuracy and uniformity. On the other hand, asynchronous updating might be more suitable for streaming data and highly distributed systems, where flexibility, rapid adaptation, and scalability are prioritized, allowing for frequent updates and quick responses to evolving data patterns. So due to the FCL environment, in which we can see dynamic client participation and incremental data, asynchronous updating might be favored to allow rapid adaptation. 

\begin{figure}[t]
  \centering
  \includegraphics[width=0.9\columnwidth ,trim={0.7cm 1.7cm 1.2cm 1.2cm},clip]{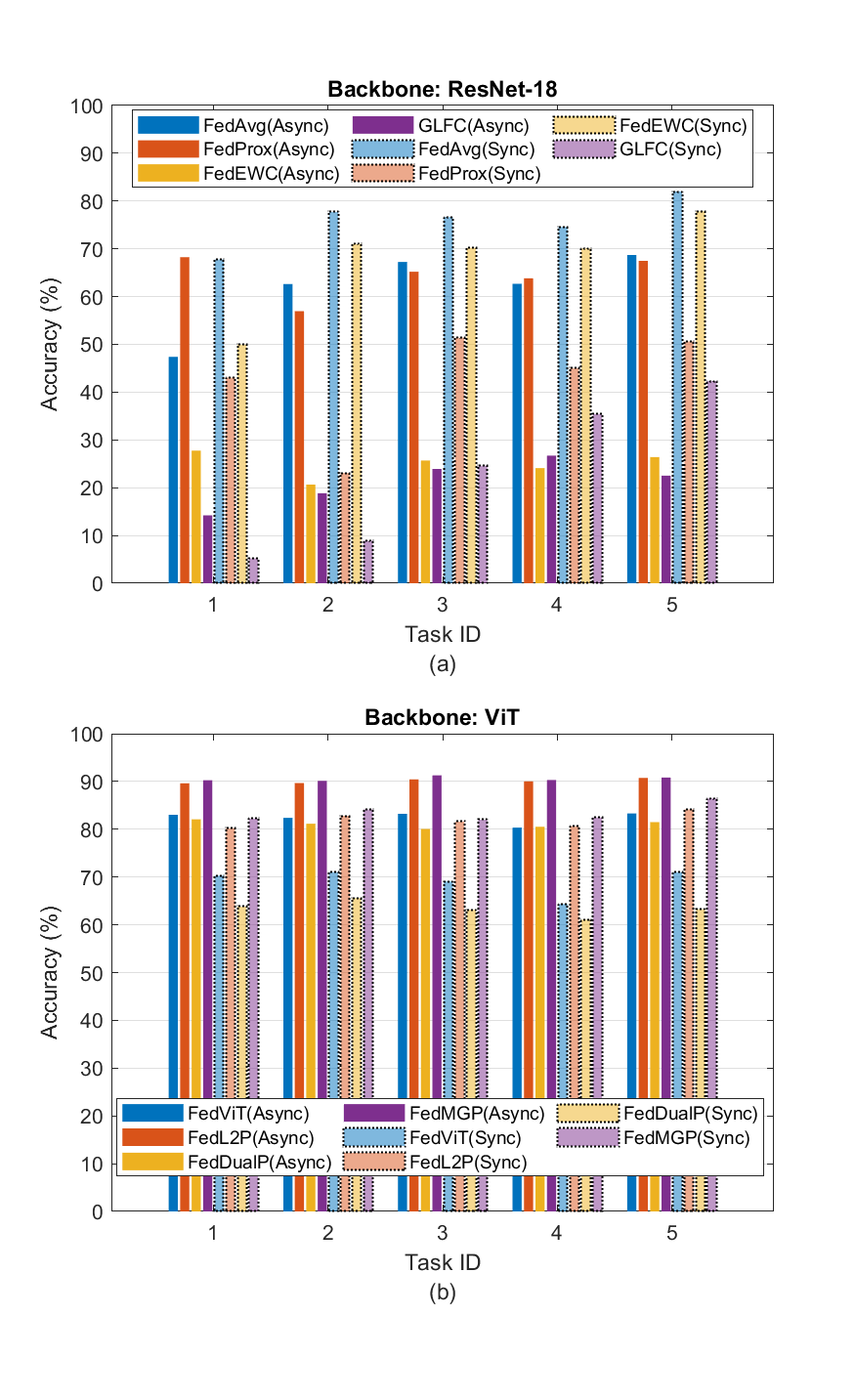}
  \caption{Evaluation of the aggregated global model's accuracy on local CIFAR-100 test sets, with each client following a 5-task class-incremental learning setup.}
  \label{fig:SyncAsyncGlobalModel}
\end{figure}

The evaluation is conducted on CIFAR-100, partitioned into 5 class-incremental tasks to simulate realistic FCL scenarios. Two FCL settings are considered: synchronous, where all clients follow the same task sequence with varying class proportions (controlled via Dirichlet distribution, $\alpha=1$), and asynchronous, where clients have partially overlapping class distributions, some classes are globally shared while others remain private to individual clients. To examine the impact of these settings on global model performance, several incremental aggregation methods are compared, including FedAvg \cite{pmlr-v54-mcmahan17a}, FedProx \cite{MLSYS2020_1f5fe839}, FedEWC \cite{doi:10.1073/pnas.1611835114}, and GLFC \cite{Dong_2022_CVPR} with a ResNet-18 backbone, as well as FedViT \cite{ZUO20241}, FedL2P \cite{Wang_2022_CVPR}, FedDualP \cite{10.1007/978-3-031-19809-0_36}, and FedMGP \cite{Personalized} with a ViT backbone. The objective is to assess how different aggregation strategies influence global model accuracy across clients in both synchronous and asynchronous FCL settings. 
As shown in Figure~\ref{fig:SyncAsyncGlobalModel}, the darker bars represent the global model's accuracy when it is updated using synchronous aggregation, while the lighter bars correspond to asynchronous updating. In the upper plot, ResNet-18 is used as the backbone model, highlighting a clear performance gap between the two aggregation strategies. Synchronous methods, such as FedAvg and FedProx, consistently achieve higher global accuracy across tasks, suggesting more stable learning and reduced forgetting. In contrast, asynchronous methods often result in lower performance, particularly in later tasks, likely due to delayed or uncoordinated client updates that disrupt the learning process. This issue is especially pronounced in methods like GLFC, where the asynchronous variant exhibits a sharp decline in accuracy, indicating significant forgetting.
In contrast, the lower plot, where a Vision Transformer (ViT) is used as the backbone, demonstrates that ViT-based models maintain strong and consistent performance under both synchronous and asynchronous settings. Methods such as FedMGP and FedDualP achieve high accuracy across all tasks and exhibit minimal signs of forgetting. This robustness is attributed to ViT’s ability to capture global dependencies through self-attention mechanisms and its strong representation learning capabilities, which help preserve previously acquired knowledge even when updates are asynchronous.
In summary, synchronous aggregation generally leads to more reliable performance and better retention of learned knowledge, particularly with convolutional architectures like ResNet-18. However, ViT-based models demonstrate that asynchronous aggregation can still perform competitively, making it a practical alternative in scenarios where full client synchronization is difficult to achieve.

In terms of communication efficiency, asynchronous updating is generally more efficient than synchronous methods, as it allows clients to send updates independently without waiting for others, reducing idle time and making better use of network resources. This is particularly advantageous in FCL, where data arrives incrementally and timely updates are crucial. In contrast, synchronous approaches suffer from communication bottlenecks due to client synchronization requirements, often leading to delays and underutilized bandwidth, especially in environments with heterogeneous client speeds or intermittent availability.

In FCL, client participation can be categorized into two types: static and dynamic. Static client participation involves a fixed set of clients that consistently contribute to the training process over time. Dynamic client participation, on the other hand, includes a variable set of clients that can join or leave the training process at different times, allowing for more flexibility and adaptability in the learning process. These two forms of participation are compared in Table \ref{tab:static}.

While static client participation offers advantages such as consistency, stability in model aggregation, predictable resource allocation, reduced communication overhead, and enhanced data privacy due to a stable client pool, it is not well-suited for the dynamic nature of FCL.
The limitations of static client participation in FCL include limited data diversity, scalability challenges, vulnerability to client dropout or failure, potential data staleness, and a lack of flexibility. Static participation may not provide the diverse and evolving datasets needed for robust model training, making it difficult to adapt to changes in data distributions or accommodate new clients. This rigid structure hinders the continuous adaptation and real-time data updates essential for effective learning in dynamic environments.
Thus, static client participation fails to meet the adaptable and scalable requirements of FCL, which thrives on dynamic client engagement and continuous data contributions to maintain model relevance and accuracy over time.

Dynamic client participation in FCL  addresses the challenges of modern environments, such as client addition and removal, limited memory, communication overhead, and increased latency due to a large number of clients \cite{9562751}. This approach allows the system to adaptively include or exclude clients during the training process, accommodating changes in client availability, connectivity, and data quality.

Dynamic participation offers several advantages, including enhanced adaptability to changing conditions, efficient resource utilization, improved privacy, and better model generalization. Although this approach introduces challenges such as increased complexity in managing clients, communication overhead, potential data imbalance, security risks, and latency issues, these can be mitigated through the use of appropriate algorithms and robust coordination mechanisms. The ability to dynamically adjust to real-world conditions and continuously enhance the model makes dynamic client participation an effective and flexible strategy for FCL in diverse and real-world applications.

\begin{table}[t]
	\centering
	\footnotesize
	\caption{Summary of advantages and disadvantages of static and dynamic client participation.}
	\label{tab:static}
	\begin{tabularx}{\columnwidth}{p{3mm} *{2}{>{\RaggedRight\arraybackslash} X }}
		\toprule
		 & Advantages & Disadvantages \\ 
		\midrule
		\multirow{5}{*}{\rotatebox[origin=c]{90}{Static}} 
		& Consistency and stability \newline Simplified model aggregation \newline Predictable resource allocation \newline Lower communication overhead \newline Data privacy and security 
		& Limited data diversity \newline Scalability challenges \newline Client dropout or failure \newline Data staleness \newline Lack of flexibility \\ 
		\midrule
		\multirow{8}{*}{\rotatebox[origin=c]{90}{Dynamic}} 
		& Improved adaptability \newline Efficient resource utilization \newline Enhanced privacy and security \newline Mitigation of stragglers  \newline Robust in real-world scenarios \newline Improved generalization \newline Continuous model improvement \newline --
		& Complex client management \newline Communication overhead \newline Data imbalance and bias \newline Security risks \newline Computational overhead \newline Latency issues \newline Dependent on accurate data \newline Complex coordination \\
		\bottomrule
	\end{tabularx}
\end{table}

\subsection{Client Models Challenges}
\begin{figure}[t]
  \centering
  \includegraphics[width=\columnwidth , page=3,trim={0.5cm 0.5cm 0.5cm 0.5cm},clip]{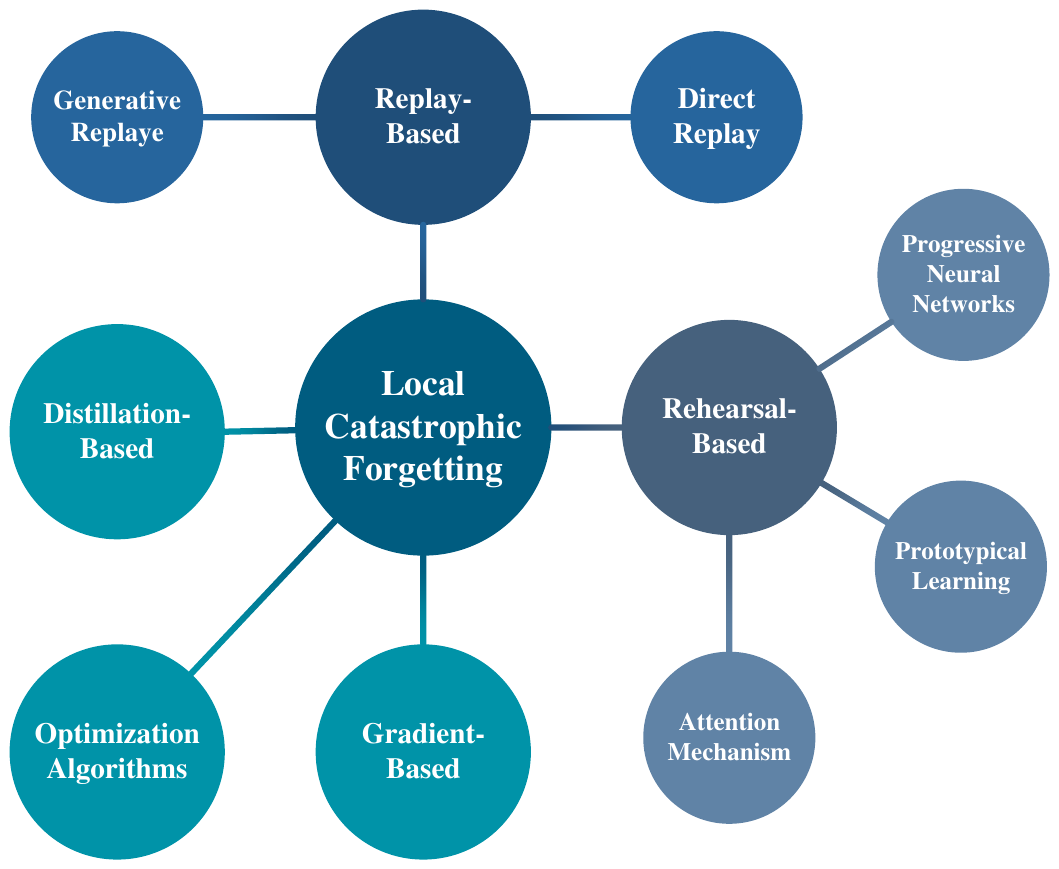}
  \caption{Taxonomy of methods used to tackle local forgetting.}
  \label{fig:local_forget}
\end{figure}

Local catastrophic forgetting refers to the phenomenon where an artificial neural network loses previously acquired knowledge about specific parts of the input space when it learns new information. This issue is particularly prevalent in sequential learning tasks, where the model updates its weights based on new data, causing interference with the knowledge acquired from earlier data. Unlike global catastrophic forgetting, which affects the overall performance on earlier tasks, local catastrophic forgetting impacts specific local regions or features within the learned tasks.

The significance of local catastrophic forgetting is profound in real-world applications, where continuous learning is required, such as autonomous driving, medical diagnosis, and robotics. Understanding and mitigating this problem is crucial for the development of robust and reliable AI systems capable of lifelong learning. As shown in Figure \ref{fig:local_forget}, researchers have explored various strategies to address local catastrophic forgetting, including regularization techniques, memory replay methods, and architectural modifications to neural networks.

Client catastrophic forgetting is another FCL challenge that has roots in both FL and CL challenges. If clients had sufficient memory and no constraints, this challenge could be mitigated. However, devices often lack adequate memory, which is a fundamental issue in FL. Additionally, when incoming data exhibits significant distribution changes (drift) or introduces new tasks (class-incremental learning), it becomes difficult for client models to retain previously learned information, while acquiring new knowledge. When incoming data is not enough, the model has more difficulty in forming strong representations of new knowledge and is, therefore, shallow and forgetful. Under this condition, memory constraints lead to overfitting, as the model ends up learning the limited available dataset instead of proper generalization. In addition, this disparity decreases the model's memory capacity for new knowledge, especially in non-IID cases, where the data is less representative of distribution shifts or new tasks.

By identifying and addressing local catastrophic forgetting, we can enhance the adaptability and efficiency of AI systems, ensuring that they retain valuable knowledge over time, while continuously learning from new experiences. The solutions are categorized into distillation-based methods, replay-based methods, rehearsal-based methods, optimization algorithms, and gradient-based methods.

Distillation-based methods involve transferring knowledge from an existing model (the teacher) to a new model (the student) in a way that helps in preserving previously learned information while accommodating new data \cite{9821057, JIN202457}. \cite{Psaltis_2023_ICCV} extracted meaningful representations and performed continual distillation locally. \cite{liu2023fedetcommunicationefficientfederatedclassincremental,10148063} utilized enhancer and replay mechanisms to balance old and new knowledge and enhance learning.

Many studies explore replay-based methods, which can be divided into direct replay and generative replay \cite{NIPS2017_0efbe980, 10540639}. These methods store a subset of past experiences or data and periodically replay them to the model, while learning new information, helping to maintain the model's performance on previously learned tasks. Direct replay involves storing actual data samples from previous tasks and interleaving these samples with new data during the training of new tasks \cite{10540639, 10097140, WANG2023551}. Some studies \cite{10128673, good2023coordinatedreplaysampleselection, Li_2024_CVPR} select the most representative samples for the replay buffer. On the other hand, generative replay involves using generative models to synthesize data samples that approximate the distribution of previous tasks. Instead of storing actual data, the method generates representative samples from the learned distribution of past tasks during the training of new tasks \cite{NEURIPS2023_d160ea01, park2021tacklingdynamicsfederatedincremental}. In the context of CL, several studies have leveraged feature representations to address the problem of catastrophic forgetting. Wang et al. \cite{pmlr-v202-wang23ar} enhance rehearsal-based methods by incorporating loss functions based on the Hilbert-Schmidt Independence Criterion (HSIC) into the standard rehearsal framework. This approach aims to mitigate forgetting by explicitly reducing inter-task interference and promoting the learning of task-invariant features, proving particularly effective when working with limited buffer sizes. Similarly, Liu et al. \cite{liu2023centroiddistancedistillationeffective} propose a method that combines centroid-based sampling with centroid distance distillation. This strategy reduces dataset bias during rehearsal and helps preserve essential inter-class relationships from previous tasks, thereby alleviating forgetting. Prototypical learning \cite{hendryx2021federatedreconnaissanceefficientdistributed}, progressive neural networks \cite{9892815}, and attention models \cite{ZHU2022168} are used in FCL frameworks to preserve these embeddings according to their importance \cite{Psaltis_2023_ICCV} and use them for retaining previously learned information.

Gradient-based methods aim to mitigate client forgetting by manipulating the gradient during local training in each client separately to ensure that updates made, while learning new tasks, do not interfere with the knowledge of previously learned tasks. For instance, Dong et al. \cite{10323204, Dong_2022_CVPR,Dong_2023_CVPR,Dong_2023_ICCV}  introduce gradient compensation loss to ensure that learning new classes does not lead to forgetting old classes. In \cite{10323204}, a new method called the Local-Global Anti-forgetting model is proposed, which enhances the performance of GLFC \cite{Dong_2022_CVPR}. Additionally, orthogonal gradient aggregation \cite{https://doi.org/10.1002/int.22727} ensures that updates for new tasks are orthogonal to those for old tasks by minimizing the overlap between gradient directions for different tasks. Furthermore, \cite{FU2023109310, Guo_2023_CVPR} explore gradient-based methods specifically in
CL to mitigate forgetting, these methods can be used as local settings for FCL to alleviate local forgetting. 
\cite{FU2023109310} proposed Knowledge Aggregation Networks (KANets), a dual-branch architecture in which one branch is frozen to preserve previously learned knowledge, while the other remains trainable to acquire new information. This structural separation mitigates gradient conflicts between new and existing knowledge, enabling a balanced trade-off between learning new data and preventing the forgetting of prior knowledge. \cite{Guo_2023_CVPR} proposed Gradient Self-Adaptation (GSA) to alliviate the imbalance between the previously learnt and new data. Also, existing class incremental learning methods often assume all old classes forget uniformly, which is inaccurate and hinders effective knowledge retention. Some papers propose optimization and loss function modifications to tackle local forgetting \cite{9349197}. Split-based methods, knowledge extraction, and local optimization are used as optimization methods. For example, specific modules designed on the client side propose loss functions such as class-projection relation strengthening loss function, class-aware distribution balancing loss function \cite{JIANG2024127956}, and proximal term loss \cite{Yuan_2023_CVPR} to alleviate forgetting from streaming data.

\subsection{Privacy Challenges of Federated Continual Learning}
FCL combines the distributed, privacy-sensitive nature of FL with the dynamic, non-stationary data handling of CL. This combination makes traditional FL challenges like heterogeneity, privacy preservation, communication overhead, and resource constraints significantly more complex in the context of continuous data streams. Furthermore, FCL introduces new challenges like longitudinal data leakage and requires addressing the interaction of FL and CL issues, such as managing catastrophic forgetting at both global and local levels and mitigating negative knowledge transfer in a distributed, dynamic setting. Highlighting the need for new methodologies that effectively combine FL's collaboration and privacy with CL's adaptation over time is essential.

In continual scenarios, clients contribute over a period of time with evolving local data, which leads to opportunities for privacy leakage due to cumulative exposure \cite{10.1007/978-3-030-92310-5_39,tobaben2025differentialprivacycontinuallearning}. This is particularly concerning when an adversary analyzes a sequence of model updates from the same client, potentially uncovering time-sensitive patterns. For instance, even when individual updates appear harmless, their progression may reveal confidential trends such as changes in user behavior, health status, or location.

Recent research has adopted different approaches to mitigate the aforementioned risks. Temporal DP builds on traditional DP by noise injection that fluctuates over time, easing the total leakage across rounds of training \cite{kiani2025differentiallyprivatefederatedlearning}. Update obfuscation and compression techniques aim to conceal sensitive trends by modifying model updates in a randomized and time-dependent manner \cite{285479}. Altering client identifying markers periodically, or scheduling participations randomly aids in diminishing the association between subsequent updates made by a specific client \cite{hasircioglu2022privacyamplificationrandomparticipation}. Additionally, privacy-aware forgetting mechanisms reduce long-term exposure by down-weighting or discarding older updates \cite{serra2024federatedcontinuallearninggoes}. These emerging solutions suggest that effective privacy preservation in FCL must consider both the distribution of data across clients and the progression of client data over time.

\section{Datasets}
Evaluating FCL systems is a complex challenge that requires careful consideration, since datasets need to align with the various challenges of different learning scenarios, e.g., incremental, task-specific, and domain-specific learning. For instance, Table \ref{tab:evaluation} lists different dataset characteristics, including task types such as instance incremental and class incremental learning, and research domains like healthcare, cybersecurity, and IoT. This table additionally highlights specific dataset features such as data format and output types that add significant value to effective evaluation. By categorizing datasets based on these metrics, researchers are able to build a better comprehension of such challenges in performing realistic and standard evaluation of FCL systems.

To be able to use these techniques in contributive systems, our data requires some features to have, which are distinguished in Table \textcolor{red}{\ref{tab:sync}}, in which data requirements are 
which can be grouped into three categories. Most of the FCL research uses different publicly available datasets across various domains to validate any particular learning strategy for their challenges, including concept drift and catastrophic forgetting. Table \ref{tab:dataset} summarizes a comprehensive list of datasets and repositories on the basis of dimensionality, data format, type of drift, and respective repositories. The table also summarizes datasets to observe gradual and abrupt drifts with their dimensional parameters in terms of pixels, channels, or features. By providing this in-depth overview, Table \ref{tab:dataset} acts as a foundational point of reference for the selection of datasets that correspond to certain FCL research objectives.

The first category relates to situations, where the initial dataset of clients exhibits different distributions in behavior or both behavior and input space, which can occur in FL. Additionally, this applies if virtual drift happens or if there is no drift related to stream data. In this situation, a sufficient amount of labeled data from all participants is required at the beginning of the training process to determine their behavior. Once their behavior is established, it cannot change, so no additional labeled data is needed. On the contrary,  the clients are allowed to present different conditional probabilities, but these probabilities will remain constant over time. In this situation, enough labeled data from all participants is required at the beginning of the training process to determine their behavior. Once their behavior is settled, it is not possible for it to change, so no more labeled data is required.

If the clients' behavior changes over time but remains consistent among the devices, then sufficient labeled data from at least one participant is required periodically. Knowing the behavior of one participant is sufficient since, in these scenarios, the behaviors of all other participants will be identical. When a real concept drift occurs, the labeled data from that client will enable the model to detect the drift and respond appropriately.

\section{Future Research Directions}
\label{sec:future}

A pressing concern in FCL is the effective management of heterogeneity, which manifests both in the statistical properties of client data and in system resources. Variability in data distributions across clients poses significant challenges, particularly in non-IID scenarios. To address this, future research should focus on dynamic domain adaptation techniques that align data distributions across clients over time, enabling seamless collaboration. Personalization remains another key area of exploration. While global models aim for generalization, client-specific or group-level personalized models could balance local and global performance, enhancing adaptability and preserving the system’s overall efficiency and privacy.

Catastrophic forgetting, a major challenge in CL, becomes even more complex in federated systems, where global and local forgetting interact. Traditional methods such as regularization and replay need to be adapted for distributed and resource-constrained environments. For instance, new regularization strategies could preserve knowledge across tasks without excessive computational overhead, while exemplar-free generative replay techniques may offer memory-efficient solutions by synthesizing past data representations instead of storing them. These methods are crucial for maintaining performance on previous tasks while adapting to new ones in decentralized learning frameworks.

FCL systems inherently operate in environments with resource limitations and significant communication overhead. Efficient use of bandwidth and computational resources is critical for their scalability. Innovative parameter compression methods and sparsity-inducing algorithms can reduce communication costs without compromising model accuracy. Furthermore, asynchronous updating mechanisms, which allow clients to transmit updates independently, can enhance scalability by accommodating dynamic participation and minimizing delays. These strategies would ensure that FCL systems remain robust and responsive even in environments with fluctuating resource availability.

Privacy preservation is a cornerstone of FL and, by extension, FCL. As FCL relies on decentralized data, robust privacy mechanisms are essential. Research should continue to refine secure aggregation techniques, enabling efficient and private communication between clients and central servers. Homomorphic encryption and differential privacy approaches must be optimized for continuous learning scenarios, balancing the trade-off between privacy and model utility. The development of lightweight privacy-preserving algorithms is particularly crucial for deployment on resource-constrained edge devices.

This survey examines the inherent challenges within FL, CL, and FCL, focusing specifically on those emerging from the learning process itself. Accordingly, future research should prioritize enhancing the robustness and generalization of FCL models, particularly in adversarial settings and scenarios with noise or unreliable data. Strengthening defense mechanisms against adversaries and improving model resilience to label noise and distribution shifts are essential for ensuring the reliability and security of FCL systems. 

The integration of multi-task and transfer learning paradigms into FCL represents a promising avenue for extending its capabilities. Cross-domain knowledge transfer can help models generalize better by leveraging shared patterns across tasks, especially in settings with sparse or incomplete data. This requires new methods to minimize negative knowledge transfer, ensuring that information from one task does not harm performance on another. Task-aware architectures, designed to differentiate representations for different tasks, could enhance scalability and reduce task interference, making FCL systems more effective in diverse and dynamic environments.

Investigating advanced dynamic optimization techniques from related domains, such as those developed for Dynamic Multiobjective Optimization Problems (DMOPs) \cite{10634796, 10197255}, holds promise for addressing persistent challenges in Federated Continual Learning (FCL). In particular, concepts from Evolutionary Transfer Optimization (ETO), which aims to leverage knowledge from previous dynamic environments to accelerate adaptation in new settings, are highly relevant \cite{10634796}. For example, incorporating methods like Cascaded Fuzzy Systems (CFS) to generate soft labels for historical information could enable more expressive representations of the global model’s retained knowledge \cite{10634796}. Similarly, techniques such as Kernel Mean Matching (KMM), which estimate the relevance of past data by aligning data distributions without requiring explicit labels, offer a principled way to assess the applicability of prior knowledge to new client data or tasks \cite{10197255}. Integrating these ideas into the FCL framework could support the development of adaptive regularization or aggregation strategies, helping to mitigate negative transfer and reduce catastrophic forgetting. Such approaches could significantly enhance the robustness and adaptability of global models operating in non-stationary, decentralized environments.

Handling concept drift, where data distributions change over time, is another critical area for FCL. Context-aware adaptation techniques that leverage external contextual information could allow models to detect and adjust to such changes dynamically. Similarly, incorporating incremental and few-shot learning methods can enable systems to learn effectively from minimal new data, maintaining their adaptability without sacrificing prior knowledge. These capabilities are particularly important in non-stationary environments, where data streams continuously and unpredictably.

The evaluation of FCL systems also requires significant advancements. Current benchmarks and metrics fail to fully capture the unique challenges of FCL, such as simultaneous global and local learning, dynamic client participation, and resource constraints. Holistic benchmarks that simulate real-world FCL scenarios, including non-IID data and class-incremental tasks, are essential. Moreover, new metrics should assess not only accuracy and forgetting but also fairness among clients and system efficiency, providing a comprehensive view of FCL performance.

Finally, the applicability of FCL extends across various domains, each with unique requirements. In personalized healthcare, FCL can enable continuous, privacy-preserving learning from patient data, while in IoT applications, it can support real-time analytics for highly dynamic environments. Edge-based autonomous systems, such as self-driving cars or smart drones, would particularly benefit from FCL’s ability to adapt to evolving data and tasks. Tailoring FCL methods to the specific challenges and constraints of these domains could significantly expand its impact.

\section{Conclusion}
\label{sec:conclusion}
This paper provides an overall survey on FCL, exploring the intersection of FL and CL for tackling dynamic, distributed, and privacy-sensitive environments. We reviewed some key challenges in FCL, such as catastrophic forgetting, concept drift, and heterogeneity of data, together with existing methodologies and solutions. The survey also brings out the role of aggregation strategies, client participation models, and learning techniques for ensuring stability and scalability in FCL systems. Furthermore, we have identified important research gaps regarding the development of more efficient resource management and privacy-preserving approaches for decentralized settings. Through this survey, we aim to provide a foundation for future works in FCL, guiding the development of adaptive, scalable, and robust models for real-world applications.

\section*{Acknowledgements}
This work is supported by the Natural Sciences and Engineering Research Council of Canada (NSERC) under funding reference numbers CGSD3-569341-2022 and RGPIN-2021-02968.

\bibliographystyle{elsarticle-num}
\bibliography{main}

\end{document}